\documentclass[graybox,11pt]{svmult}

\usepackage{times,epsfig,graphicx}
\usepackage{amsmath,amssymb,amsopn,amsfonts}
\usepackage{algpseudocode}
\usepackage[ruled,noend]{algorithm2e}

\algnewcommand\algorithmicinput{\textbf{Input:}}
\algnewcommand\Input{\item[\algorithmicinput]}
\algnewcommand\algorithmicoutput{\textbf{Output:}}
\algnewcommand\Output{\item[\algorithmicoutput]}

\usepackage[latin1]{inputenc}
\usepackage{fancyhdr}
\usepackage{verbatim}
\usepackage{tabularx}
\usepackage{hyperref}
\usepackage{xspace}
\usepackage{bm}

\makeatletter

\DeclareRobustCommand\onedot{\futurelet\@let@token\@onedot}
\def\@onedot{\ifx\@let@token.\else.\null\fi\xspace}

\def\eg{\emph{e.g}\onedot} 
\def\ie{\emph{i.e}\onedot} 
 
\def\etc{\emph{etc}\onedot} \def\vs{\emph{vs}\onedot} 
 
\def\etal{\emph{et al}\onedot}

\newcommand{\calD}{{\cal D}}
\newcommand{\calY}{{\cal Y}}
\newcommand{\calX}{{\cal X}}
\newcommand{\calT}{{\cal T}}

\newcommand{\bfC}{\mathbf{C}}

\newcommand{\bfX}{{\bf X}}
\newcommand{\bfY}{{\bf Y}}
\newcommand{\bfW}{{\bf W}}
\newcommand{\bfS}{{\bf S}}

\newcommand{\bfP}{{\bf P}}
\newcommand{\bfQ}{{\bf Q}}
\newcommand{\bfx}{{\bf x}}

\newcommand{\eye}{\mathbf{I}}

\newcommand{\bfzero}{{\bf 0}}

\begin{document}

\title*{Domain Adaptation for Visual Applications: A Comprehensive Survey}

\author{{\Large Gabriela Csurka} }


\institute{Xerox Research Center Europe (\url{www.xrce.xerox.com}),  
6 chemin Maupertuis, 38240 Meylan, France, 
\email{Gabriela.Csurka@xrce.xerox.com}}

\authorrunning{Gabriela Csurka}

%
%
%
%

\maketitle

\abstract{{\em The aim of this paper\footnote{Book chapter to appear in "Domain Adaptation in Computer Vision Applications", 
Springer Series: Advances in Computer Vision and Pattern Recognition, Edited by Gabriela Csurka.}
 is to give an overview of domain adaptation and transfer learning 
with a specific view on visual applications. After a general motivation, we first position 
domain adaptation in the larger transfer learning problem. 
Second, we try to address and analyze briefly the state-of-the-art 
methods for different types of scenarios, first describing the historical shallow methods, addressing both the 
homogeneous and the heterogeneous domain adaptation methods. Third, we discuss the effect of
 the success of  deep convolutional architectures which led  to new type of domain adaptation  
methods that  integrate the adaptation  within the deep architecture.  
Fourth, we overview the methods that go beyond image categorization, 
such as object detection or image segmentation, video analyses 
or learning visual attributes. Finally, we conclude the paper with a section where we relate 
domain adaptation to other machine learning solutions.}}

\section{Introduction}
\label{sec:intro}

While huge volumes of unlabeled data are generated and made available in many domains, 
the cost of acquiring data labels remains high. To overcome the burden of annotation, alternative  solutions  
have been proposed in the literature  in order to exploit the unlabeled data  
(referred  to as semi-supervised learning),  
or data and/or models  available in similar domains (referred to as transfer learning). 
Domain Adaptation (DA) is a particular case of transfer learning (TL) that leverages labeled
 data in one or more related {\it source} domains, 
to learn a classifier for unseen or unlabeled data in a {\it target} domain. In general it is assumed that the 
task is  the same, \ie class labels are shared between domains.  
The source domains are assumed to be related to the target domain, but not identical, in which case, it 
becomes a standard machine learning (ML) problem where we assume that 
 the test data is drawn from the same distribution as the  training data.  When this assumption is not verified, 
\ie  the distributions of training and test sets do not match, the performance at test time can be 
 significantly degraded.   

In visual applications,  such distribution difference, called domain shift,   are common  in real-life applications.  
They can be consequences of changing conditions, \ie background, location, pose changes,  but the domain mismatch
might be more severe when, for example, the source and target domains contain  images of 
different types, such as  photos, NIR images, paintings or sketches \cite{KlareICB12STowards,CrowleyCVAA14Search,CastrejonCVPR16Learning,SaxenaTASKCV16Heterogeneous}. 
Service provider companies are especially concerned since,
for the same service (task), the distribution of the data may  vary a lot  from one customer to another. 
In general,  machine learning components of service solutions that are re-deployed from a given customer or location to
 a new customer or location require specific customization to accommodate the  new conditions. 
 For example,  in brand sentiment management it is critical to 
 tune the models to the way users  talk about their experience given  the different products.   In  
surveillance and urban traffic understanding,  pretrained  models on previous locations might need adjustment 
to the new environment. All these entail either acquisition of annotated data in the new field  
or the calibration of the pretrained models to achieve the contractual performance in the new situation. 
However, the former solution, \ie data labeling, 
is expensive and time consuming due to the significant amount of human effort involved. Therefore, 
 the second option is preferred  when possible. This can be achieved  either by adapting  the
 pretrained  models taking advantage of the unlabeled (and if available labeled) target set or, 
 to  build the target model, by exploiting  both previously acquired labeled source data  and the new unlabeled target data  together.

\begin{figure}[ttt]
\begin{center}
\vspace{0.5cm}
\includegraphics[width=0.9\textwidth]{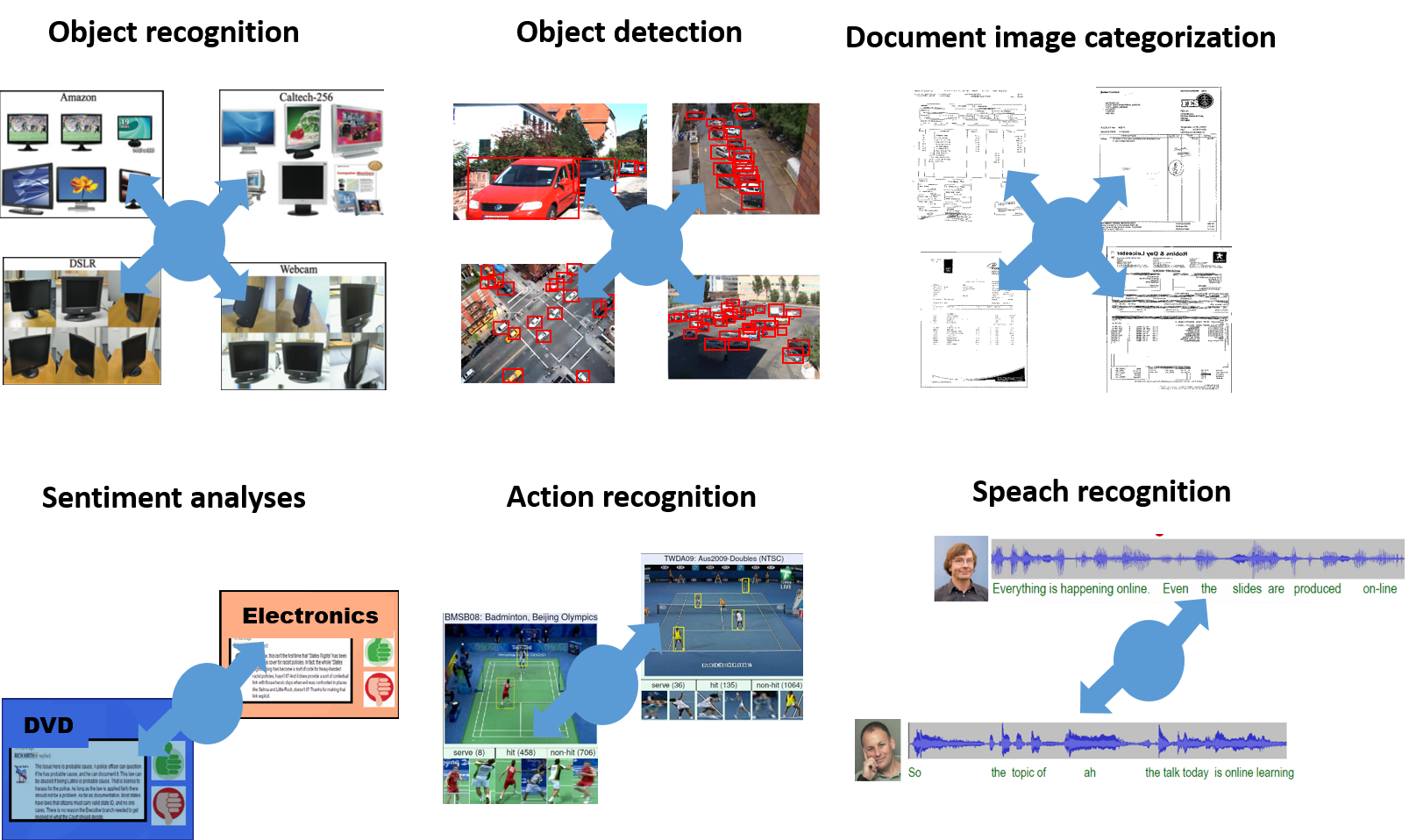}
\vspace{0.5cm}
\caption{Example scenarios with domain adaptation needs.} 
\label{fig.DAexamples}
\end{center}
\end{figure}

 Numerous approaches have been proposed in the last years to address adaptation needs  that arise in 
different application scenarios (see a few examples in Figure \ref{fig.DAexamples}). 
Examples include DA and TL solutions for named entity recognition and opinion extraction across different text corpora 
\cite{DaumeJAIR06Domain,PanWWW10Crossdomain,BlitzerAISTATS11Domain,ZhouSIGKDD14Unifying}, multilingual text 
classification \cite{PrettenhoferACL10Crosslanguage,ZhouAISTATS14Heterogeneous,ZhouAAAI14Hybrid},
sentiment analysis \cite{ChenICML12Marginalized,GlorotICML11Domain},  WiFi-based localization \cite{PanTNN11Domain},
speech recognition across different speakers \cite{LeggetterCSL95Maximum,ReynoldsDSP00Speaker},
object recognition in images acquired in different 
conditions \cite{SaenkoECCV10Adapting,GongCVPR12Geodesic,FernandoICCV13Unsupervised,LongICCV13Transfer,SunAAAI16Return},  
video concept detection \cite{YangICCV13Crossdomain}, video  event recognition \cite{DuanCVPR12Exploiting},
 activity recognition \cite{FarajidavarBMVC12Domain,ZhuBMVC13Enhancing}, human motion parsing from videos \cite{HaoquanECCV14CUnsupervised},
face recognition \cite{YangICCV11Fisher,ShekharCVPR13Generalized,SharmaCVPR11Bypassing}, 
facial landmark localization \cite{SmithECCV14Collaborative}, 
facial action unit detection \cite{ChuCVPR13Selective}, 3D pose estimation \cite{YamadaECCV12OBias},  
document categorization across different customer 
datasets \cite{ChidlovskiiKDD16Domain,CsurkaTASKCV15Adapted,CsurkaX16What}, \etc.

\begin{figure}[ttt]
\begin{center}
\vspace{0.2cm}
\includegraphics[width=.9\textwidth]{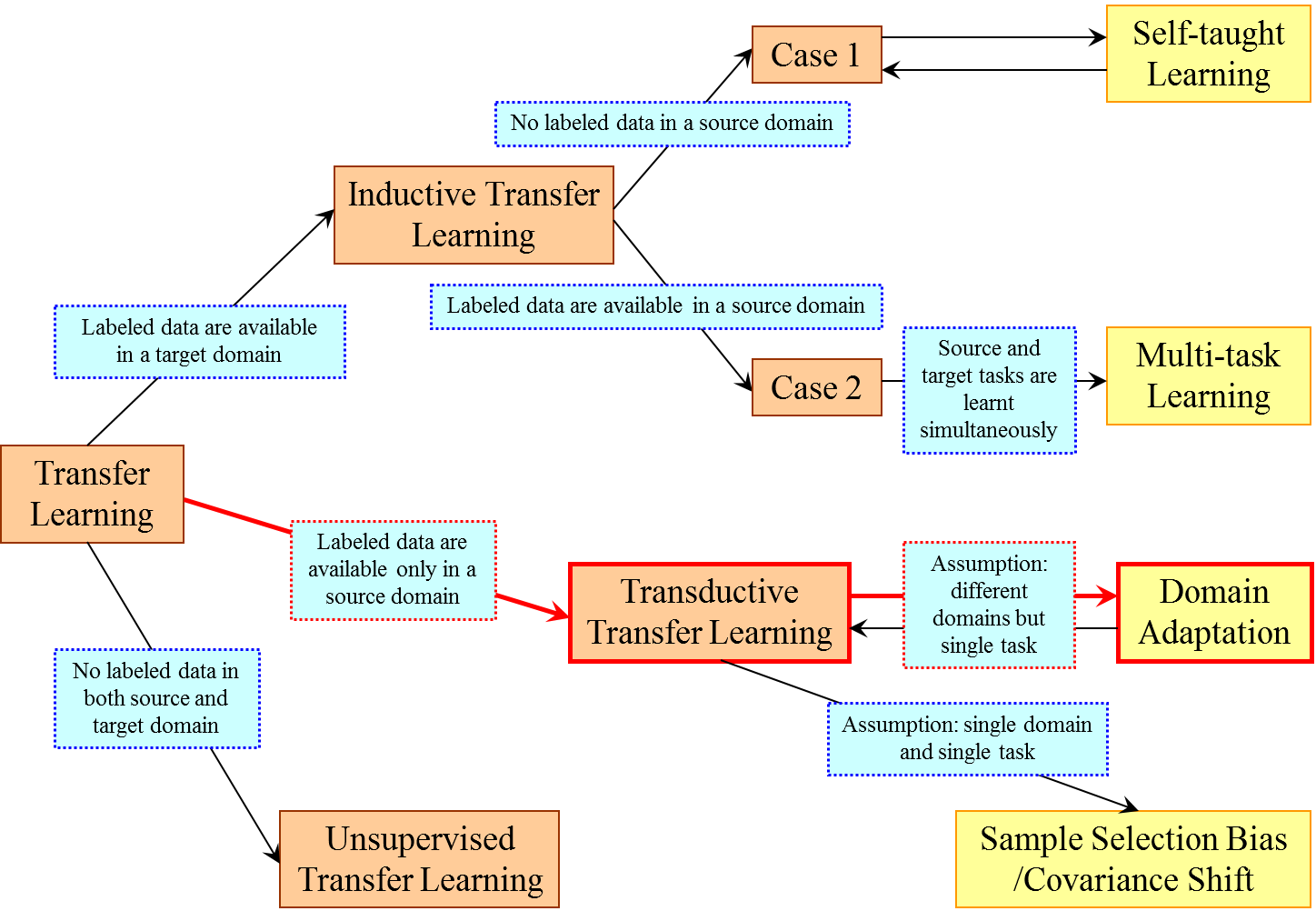}
\vspace{0.2cm}
\caption{An overview of different transfer learning approaches. (Image: Courtesy to S.J. Pan~\cite{PanTKDE10Survey}.)} 
\label{fig:TL}
\end{center}
\end{figure}

In this paper,  we mainly focus  on {\em domain adaptation} methods 
applied  to {\em visual tasks}. For a broader review of the transfer learning  literature 
as well as for approaches specifically designed to solve non-visual tasks, \eg text or speech, 
please refer to \cite{WeissJBD16Survey}. \\

The rest of the paper is organized as follows.  In Section \ref{sec.definitions}  we  define  more formally  
transfer learning  and domain adaptation. 
In Section \ref{sec.shallow}  we review shallow DA methods that can be applied on visual 
features extracted from the images, both in the homogeneous  and 
heterogeneous case.   Section \ref{sec.deep} addresses more recent deep DA methods and
Section \ref{sec:beyondclassif} describes DA solutions proposed 
for computer vision  applications  beyond image classification.  In
Section~\ref{sec:discussion} we relate  DA  to other transfer learning  and  
standard machine learning  approaches and in
 Section~\ref{sec:conclusion} we conclude the paper.

\section{Transfer learning and domain adaptation}
\label{sec.definitions}

In this section, we follow the definitions and notation of \cite{PanTKDE10Survey,WeissJBD16Survey}. 
Accordingly, a domain $\calD$ is composed of a $d$-dimensional feature space $\calX \subset R^d$  with 
a marginal probability distribution $P(\bfX)$,  and a task $\calT$  defined 
 by a label space $\calY$ and the conditional probability distribution $P(\bfY|\bfX)$,  
 where $\bfX$ and $\bfY$ are random variables. 
 Given a particular sample set  $\bfX=\{\bfx_1,\ldots \bfx_n\}$ of $\calX$, 
 with corresponding labels  $\bfY=\{y_1,\ldots y_n\}$ from $\calY$, 
$P(\bfY|\bfX)$ can  in general  be learned in a supervised manner 
from these feature-label pairs  $\{\bfx_i, y_i\}$.

Let us assume that we have two domains with their related tasks:
a {\em source}  domain $\calD^s=\{\calX^s,P(\bfX^s)\}$ with $\calT^s=\{\calY^s,P(\bfY^s|\bfX^s)\}$ and 
a {\em target} domain  $\calD^t=\{\calX^t,P(\bfX^t)\}$ with $\calT^t=\{\calY^t,P(\bfY^t|\bfX^t)\}$.
If the two domains corresponds, \ie  $\calD^s = \calD^t$ and $\calT^s =\calT^t$,   traditional ML methods 
can be used to solve the problem,  where  $\calD^s$ becomes the training set and $\calD^t$ the test set. 

When this assumption does not hold, \ie $\calD^t\neq \calD^s$ or $\calT^t\neq \calT^s$, 
the models trained on $\calD^s$ might perform poorly on  $\calD^t$, or they are
  not  applicable directly if $\calT^t\neq \calT^s$.  
When the source domain is somewhat related to the target, it is possible to exploit  the
related information from $\{\calD^s,\calT^s\}$ to learn  $P(\bfY^t|\bfX^t)$. This
 process  is known as {\em transfer learning} (TL).  
 
 We distinguish  between {\em homogeneous  TL},   where the source and  target are represented in the same 
the feature space,  $\calX^t = \calX^s$,  with  $P(\bfX^t)\neq P(\bfX^s)$  due to domain shift, and 
 {\em heterogeneous TL} where the source and  target  
 data can have different representations, $\calX^t \neq  \calX^s$ (or they can even  be of different modalities
such as image \vs text).

Based on these definitions, \cite{PanTKDE10Survey} categorizes the
TL approaches  into three main groups depending on the different situations concerning
source and target domains and the corresponding tasks. These are the  inductive TL, transductive TL
and unsupervised TL (see Figure \ref{fig:TL}).   
The {\em inductive TL} is the case where the target task
is different but related to the source task, no matter whether the
source and target domains are the same or not. It requires  at least some labeled target instances   
to induce a predictive model  for  the target data. In the case of  the {\em transductive TL},
 the source and target tasks are the same, and either the source and target data representations are different 
 ($\calX^t \neq \calX^s$)  or the source and target distributions are different due to  selection bias  
 or distribution mismatch.  Finally, the {\em unsupervised TL} refers to the case 
where  both the domains and the tasks are different but somewhat related. In general,  labels are not available 
neither for the source nor for the target, and the focus is  on exploiting the  (unlabeled) 
information in the source domain to solve  unsupervised learning 
task in the target domain. These tasks include  clustering, dimensionality reduction and
density estimation \cite{DaiICML08Selftaught,WhangPKDD08Transferred}.

According to this classification, DA methods are transductive TL solutions, where it is 
assumed that the tasks are the same, \ie $\calT^t = \calT^s$.  In general they refer to a categorization task, where 
 both the set of  labels  and the
conditional distributions  are assumed to be shared between the two domains, \ie $\calY^s=\calY^t$ and 
$P(\bfY|\bfX^t) = P(\bfY|\bfX^s)$. However,  the  second  assumption  is rather  strong and does not always 
hold in real-life applications.  Therefore, the definition of domain adaptation is relaxed to 
the case where only the first assumption is required, \ie  $\calY^s=\calY^t=\calY$. 

In the DA community, we further distinguish between  the {\em unsupervised}\footnote{Note also that the 
unsupervised DA is not related to the unsupervised TL, for which  no source labels are available and 
in general the task to be solved is unsupervised.} (US) case where the
labels are available only for the source domain and the {\em semi-supervised} (SS) case where 
a small set of target examples are labeled.

\section{Shallow domain adaptation methods}
\label{sec.shallow}

In this section, we review shallow DA methods that can be applied on vectorial visual 
features extracted from  images. First, in Section \ref{sec.homogeneous}  
we survey homogeneous DA  methods, where  the feature representation for the source and  
target domains is the same,  $\calX^t = \calX^s$ with  $P(\bfX^t)\neq P(\bfX^s)$,  and 
the tasks shared,  $\calY^s=\calY^t$. 
Then, in Section \ref{sec.multisource} we discuss  methods that can exploit efficiently several  
source domains. Finally in Section~\ref{sec.heterogeneous}  we  discuss the heterogeneous case, 
where the source and  target  data have different representations.

\subsection{Homogeneous domain adaptation methods}
\label{sec.homogeneous}

\begin{figure}[ttt]
\begin{center}
\includegraphics[width=.8\textwidth]{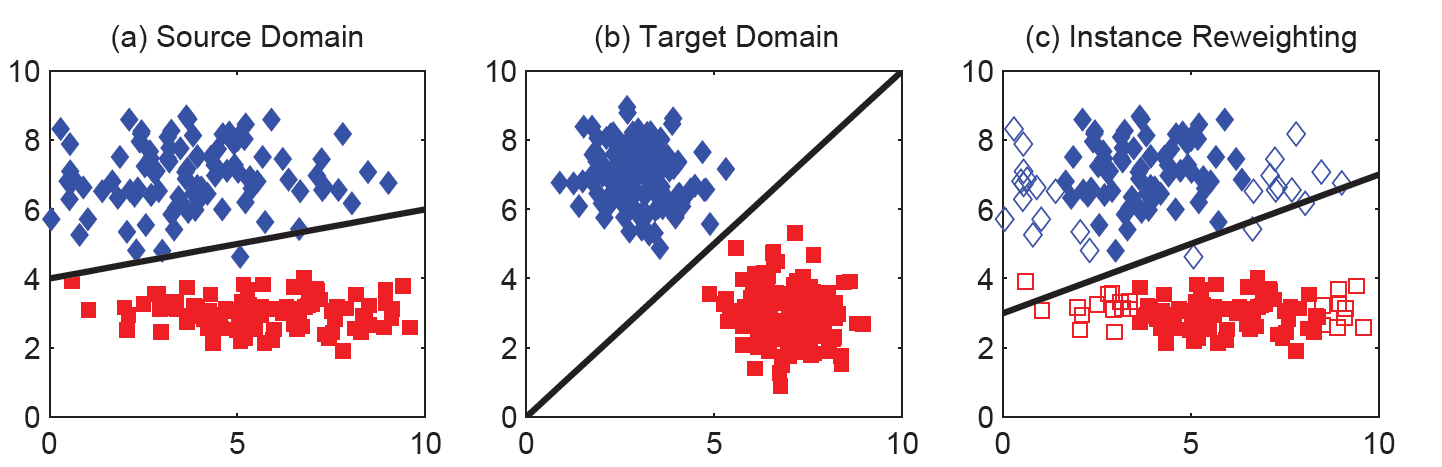}
\caption{ Illustration of the effect of instance re-weighting samples on the source classifier.
(Image: Courtesy to M. Long~\cite{LongCVPR14Transfer}.)} 
\label{fig.TJM}
\end{center}
\end{figure}

\noindent{\bf Instance re-weighting methods.}
The DA case when  we assume that the
conditional distributions  are  shared between the two domains, \ie $P(\bfY|\bfX^s) = P(\bfY|\bfX^t)$, 
is often referred to as  {\em dataset bias} or {\em covariate shift} \cite{ShimodairaJSPI00Improving}. 
In this case,  one could simply  apply  the model learned on the source to estimate  $P(\bfY|\bfX^t)$. However, as 
$P(\bfX^s)\neq P(\bfX^t)$, the source  model might  yield a poor performance when applied on 
the target set despite of the  underlying  $P(\bfY|\bfX^s) = P(\bfY|\bfX^t)$ assumption. 
The most popular early solutions proposed to overcome this to happen  are  based on  instance re-weighting
(see Figure~\ref{fig.TJM} for an illustration).

To compute the weight of an instance, early methods proposed to estimate the ratio 
between the likelihoods of being a source or target example. This can be done either by estimating 
the likelihoods independently using a domain classifier \cite{ZadroznyICML04Learning} 
or by approximating directly the ratio between the densities
with a Kullback-Leibler Importance Estimation Procedure \cite{SugiyamaNIP08Direct,KanamoriJMLR09Efficient}. 
However, one of the most popular measure used to weight data instances, 
used for example in \cite{HuangNIPS07Correcting,GrettonBC09Covariate,PanTNN11Domain},
is the Maximum Mean Discrepancy (MMD) \cite{Borgwardt06Integrating} computed between 
the data distributions in the two domains.

The method proposed in \cite{DudikNIPS05Correcting}  infers re-sampling weights through maximum entropy
density estimation. \cite{ShimodairaJSPI00Improving} improves predictive inference under covariate shift 
by weighting the log-likelihood function. The Importance Weighted Twin Gaussian Processes 
 \cite{YamadaECCV12OBias}  directly learns  the importance weight function, without going through
density estimation, by using the relative unconstrained least-squares importance fitting.
The Selective Transfer Machine \cite{ChuCVPR13Selective} jointly optimizes the weights as well as the classifier's
parameters to preserve the  discriminative power of the new decision boundary.

\begin{figure}[ttt]
\begin{center}
\includegraphics[width=.65\textwidth]{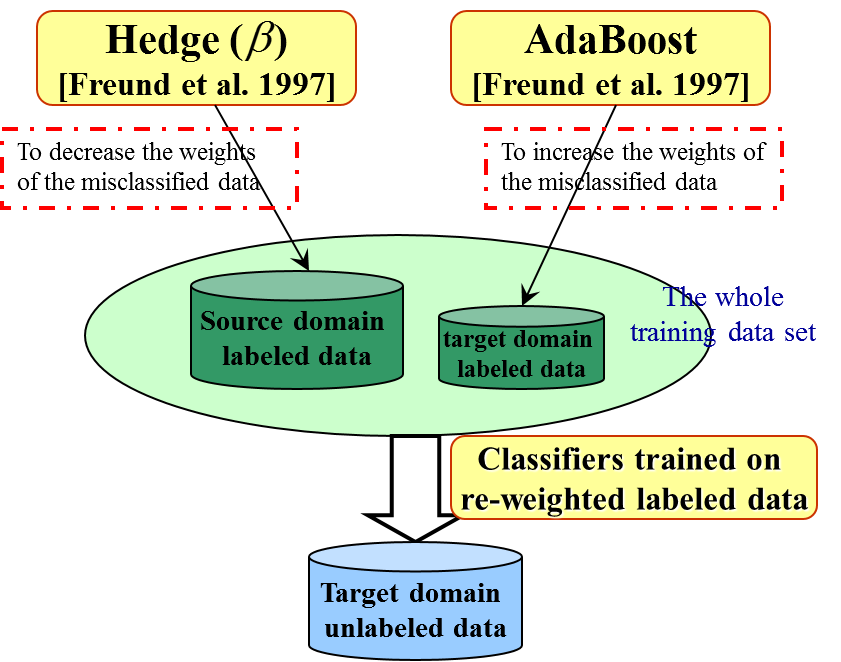}
\caption{Illustration of the TrAdaBoost method~\cite{DaiICML07Boosting} where the idea is to decrease the 
importance of the misclassified  source examples while focusing, as in AdaBoost~\cite{FreundJCSS97Decision},
on the misclassified target examples. (Image: Courtesy to S. J. Pan).} 
\label{fig.Trboost}
\end{center}
\end{figure}

 The Transfer Adaptive Boosting (TrAdaBoost) \cite{DaiICML07Boosting},  is  an extension to 
 AdaBoost\footnote{Code at \url{https://github.com/BoChen90/machine-learning-matlab/blob/master/TrAdaBoost.m}} 
 \cite{FreundJCSS97Decision}, that iteratively re-weights both source and target examples 
 during the learning of a target classifier.  This is done by increasing the weights of miss-classified target instances 
as in the traditional AdaBoost, but decreasing the weights of  miss-classified source samples in 
order to diminish their importance  during the training process (see Figure~\ref{fig.Trboost}).
 The  TrAdaBoost was  further extended  by integrating dynamic 
 updates in \cite{AlstouhiPKDD11Adaptive,ChidlovskiiCLEFWN14Assembling}.\\

\begin{figure}[ttt]
\begin{center}
\includegraphics[width=1\textwidth]{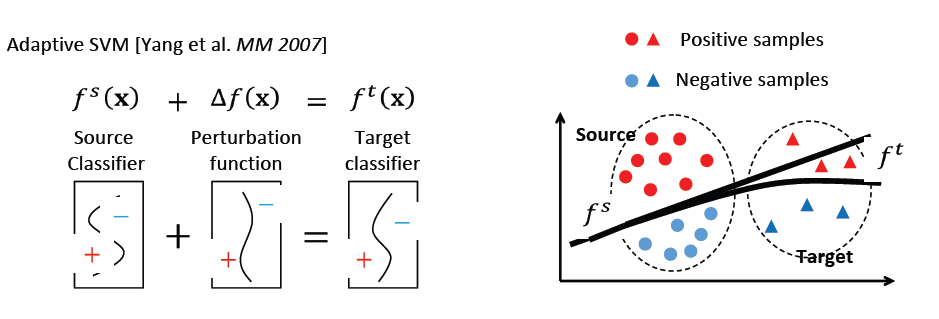}
\caption{Illustration of the  Adaptive SVM~\cite{YangMM07Crossdomain}, 
	where a set of so called perturbation functions  $\Delta_f$ are added to the 
source classifier $f^s$ to progressively  adjusts the decision boundaries of $f^s$
 for the target domain. (Courtesy to D. Xu).} 
\label{fig.asvm}
\end{center}
\end{figure}

\noindent{\bf Parameter adaptation methods.} Another set of early DA methods, 
but which does not necessarily assume 
 $P(\bfY|\bfX^s) = P(\bfY|\bfX^t)$, investigates different options to adapt
 the classifier trained on the source domain, \eg an SVM,  in order to perform 
better on the target domain\footnote{The code for several methods, such as A-SVM, A-MKL, DT-MKL 
can be downloaded from  \url{http://www.codeforge.com/article/248440}}.   
Note that these methods in general require at least a small set of  
labeled target examples per class, hence they can only be applied in the semi-supervised DA  scenario.
As such, the Transductive SVM~\cite{JoachimsICML99Transductive}  that aims at decreasing
the  generalization error of the classification,  by incorporating knowledge about the target 
 data into the SVM optimization process. The  Adaptive SVM (A-SVM)~\cite{YangMM07Crossdomain}  progressively 
 adjusts the decision boundaries of the source classifiers 
 with the help of a set of so called perturbation functions built by exploiting predictions on the available 
labeled  target examples (see Figure~\ref{fig.asvm}). 
The Domain Transfer SVM~\cite{DuanCVPR09Domain}  simultaneously
reduces the mismatch in the distributions (MMD) between
two domains and learns a target decision function. 
The  Adaptive Multiple Kernel Learning (A-MKL)~\cite{DuanCVPR12Exploiting} generalizes this by 
learning an adapted classifier based on
multiple base kernels and the pre-trained average classifier.  The model 
minimizes jointly the structural risk functional and
the mismatch between the data distributions (MMD) of the two domains.

The  domain adaptation SVM (DASVM)~\cite{BruzzonePAMI10Domain} exploits within the semi-supervised  DA scenario
both the transductive SVM~\cite{JoachimsICML99Transductive} 
and its extension, the progressive transductive SVM~\cite{ChenPRL03Learning}.
The cross-domain SVM,  proposed  in ~\cite{JiangICIP08Crossdomain},   constrains 
the impact of source data to the k-nearest neighbors (similarly to the spirit of the 
Localized SVM~\cite{ChengSICDM07Localized}). This is done by down-weighting 
support vectors from the source data that are far from the target  samples.\\

\noindent{\bf Feature augmentation.}
One of the  simplest method for DA was  proposed 
in \cite{DaumeX09Frustratingly}, where the original representation $\bfx$ is augmented with 
itself and a vector of the same size filled with zeros as follows:  the   source features become 
{\tiny $\Bigg[\begin{array}{c} \bfx^s \\ \bfx^s \\ \bfzero \end{array} \Bigg]$}
and target features  {\tiny $\Bigg[ \begin{array}{c}\bfx^t \\ \bfzero \\ \bfx^t \end{array}\Bigg]$}. 
Then an SVM is trained on these augmented features 
to figure out which parts of the representation is shared between the 
domains and which are the domain specific ones.

\begin{figure}[ttt]
\begin{center}
\includegraphics[width=1\textwidth]{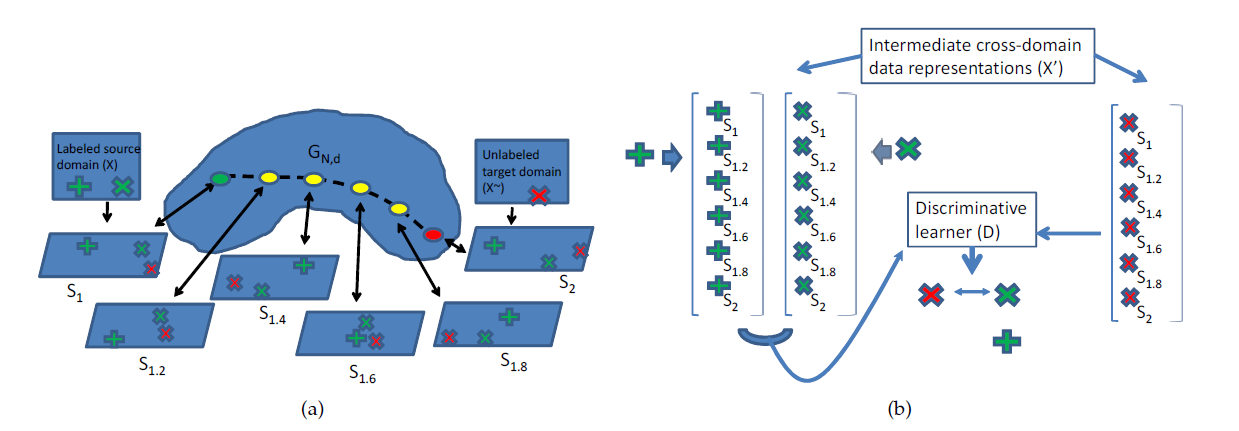}
\caption{The GFS samples   between source $S_1$ and target $S_2$ on 
the geodesic path intermediate domains $S_{1,i}$ that can be seen as cross-domain data representations.
 (Courtesy to R. Gopalan~\cite{GopalanICCV11Domain}).)} 
\label{fig.GFS}
\end{center}
\end{figure}

The idea of feature augmentation is also behind the  Geodesic Flow Sampling (GFS)
\cite{GopalanICCV11Domain,GopalanPAMI14Unsupervised}   and the
Geodesic Flow Kernel (GFK)~\cite{GongCVPR12Geodesic,GongICML13Connecting}, 
where the domains are embedded in  $d$-dimensional linear subspaces 
that can be seen as points on the Grassman manifold corresponding to the
collection of all $d$-dimensional subspaces.  In the case of GFS~\cite{GopalanICCV11Domain,GopalanPAMI14Unsupervised}, 
following the geodesic path between the source and 
target domains, representations,  corresponding to  intermediate domains,  are sampled gradually
and concatenated  (see illustration in Figure~\ref{fig.GFS}).  
Instead of sampling,  GFK\footnote{Code available at \url{http://www-scf.usc.edu/~boqinggo/domain_adaptation/GFK_v1.zip}}
\cite{GongCVPR12Geodesic,GongICML13Connecting}, 
extends GFS  to the infinite case, proposing a kernel that  makes the solution  
equivalent to integrating over all common subspaces  lying on the geodesic path.

A more generic framework, proposed  in \cite{GopalanPAMI14Unsupervised}, 
accommodates domain representations in 
high-dimensional Reproducing Kernel Hilbert Space (RKHS) using kernel methods  and low-dimensional
manifold representations corresponding to  Laplacian Eigenmaps. 
The  approach described in \cite{NiICCV2013Subspace} was 
inspired  by the manifold-based  incremental learning framework in \cite{GopalanICCV11Domain}. It
generates a set of intermediate dictionaries which smoothly connect the source and target domains. 
This is done by decomposing the target data with the current intermediate
domain dictionary updated with  a reconstruction residue  estimated  on the target.
Concatenating these intermediate representations 
enables learning a better cross domain classifier. \\

These  methods exploit {\em intermediate cross-domain} representations that are built without the use of  class labels. 
Hence,  they  can be applied in  both, the US and SS,  scenarios. These cross-domain representations are then used 
either to train a discriminative classifier~\cite{GopalanPAMI14Unsupervised} using 
the  available labeled set (only from the source or from both  domains), 
or to label the target instances using nearest neighbor search in the 
kernel space~\cite{GongCVPR12Geodesic,GongICML13Connecting}. \\

\noindent {\bf Feature space alignment.} Instead of of augmenting the features, other methods 
tries to  align  the source features with the target ones. As such, 
the Subspace Alignment (SA)~\cite{FernandoICCV13Unsupervised}  
learns an alignment between the source subspace obtained by PCA and the target 
PCA subspace, where the PCA dimensions are selected by  minimizing the Bregman  
divergence between the subspaces. It advantage is its simplicity, as shown in 
Algorithm~\ref{SA_algo:da}.  Similarly, 
the linear Correlation Alignment (CORAL)~\cite{SunAAAI16Return} 
can be written in few lines of MATLAB code as illustrated in Algorithm~\ref{coral_alg:coral}.
The method  minimizes the domain shift by
using the second-order statistics of the source and target
distributions. The main idea is a whitening of the source data using  its  covariance 
followed by a "{\em re-coloring}" using the target covariance matrix.  \\

As an alternative to feature alignment, a large set of feature transformation methods were proposed with the
objective to find a projection of the data into a latent space such that the discrepancy between the source and target 
distributions is decreased.  Note that the 
projections   can be shared between the domains  or they can be domain specific projections. In the latter case 
 we talk about asymmetric feature transformation. Furthermore, 
 when the transformation  learning  procedure uses no class labels, the method is called 
  {\em unsupervised} feature transformation and when the transformation is learned by  exploiting class labels  
(only from the source or also from the target when available) it is referred to as 
{\em supervised}  feature transformation.  \\

\begin{algorithm}[t]
\caption{\small Subspace Alignment (SA)~\cite{FernandoICCV13Unsupervised}}
\label{SA_algo:da}
\begin{algorithmic}[1]
\Input{Source data $\bfX^s$, target data $\bfX^t$, subspace dimension $d$}
\State{$\bfP_s \leftarrow PCA(\bfX^s,d)$, \,\,\, $\bfP_t \leftarrow PCA(\bfX^t,d)$  \;}
\State{$\bfX^s_a = \bfX^s \bfP_s \bfP_s^{\top} \bfP_t$, \,\,\,  $\bfX^t_a = \bfX^t \bfP_t$ \;}
\Output{Aligned source, $\bfX^s_a$ and target, $\bfX^t_a$ data.}
\end{algorithmic}
\end{algorithm}

\begin{algorithm}[t]
\caption{\small Correlation Alignment (CORAL)~\cite{SunAAAI16Return}}
\begin{small}
\begin{algorithmic}[1]
\Input{Source data $\bfX^s$, target data $\bfX^t$}
\State {$\bfC_s = cov(\bfX^s) + eye(size(\bfX^s, 2))$, \,\,\, $\bfC_t = cov(\bfX^t) + eye(size(\bfX^t, 2))$}
\State {$\bfX^s_w = \bfX^s*\bfC_s^{-1/2}$ (whitening), \,\,\, $\bfX^s_a = \bfX^s_w*\bfC_t^{-1/2}$ 
(re-coloring)}
\Output{Source data $\bfX^s_a$ adjusted to the target.}
\end{algorithmic} 
\end{small}
\label{coral_alg:coral}
\end{algorithm}

\noindent {\bf Unsupervised feature transformation.}   One of the first such  DA method is the 
Transfer Component  Analysis (TCA) \cite{PanTNN11Domain} 
that proposes to discover common latent features 
 having the same marginal distribution across the source and target domains,  
while maintaining the intrinsic structure (local geometry of the data manifold)
of the original domain by a smoothness term.

Instead of restricting the discrepancy to a simple distance between the sample means in
the lower-dimensional space, Baktashmotlagh \etal~\cite{BaktashmotlaghICCV13Unsupervised} propose the Domain 
Invariant 
Projection\footnote{Code at \url{https://drive.google.com/uc?export=download&id=0B9_PW9TCpxT0c292bWlRaWtXRHc}}
 (DIP) approach  that compares directly  the distributions in the RKHS while  
constraining the transformation to be orthogonal. 
They go a step further in \cite{BaktashmotlaghCVPR14Domain}
and based on the fact that 
probability distributions lie on a Riemannian manifold,  propose  the Statistically Invariant 
Embedding\footnote{Code at \url{https://drive.google.com/uc?export=download&id=0B9_PW9TCpxT0SEdMQ1pCNzdZekU}} 
(SIE) that uses the Hellinger distance on this manifold 
to compare kernel density estimates between of the source and target data.
Both the DIP and SIE,  involve non-linear optimizations and are solved 
with the  conjugate gradient algorithm ~\cite{EdelmanSIMAX98Geometry}.

The Transfer Sparse 
Coding\footnote{Code at \url{http://ise.thss.tsinghua.edu.cn/~mlong/doc/transfer-sparse-coding-cvpr13.zip}} 
(TSC)~\cite{LongCVPR13Transfer} learns robust sparse representations for classifying
cross-domain data accurately. To bring the domains closer, 
the distances between the sample means for each dimensions  
of the source and the target is incorporated into the objective function to be minimized.  
The Transfer Joint 
Matching\footnote{Code at \url{http://ise.thss.tsinghua.edu.cn/~mlong/doc/transfer-joint-matching-cvpr14.zip}} 
(TJM)~\cite{LongCVPR14Transfer}  learns a non-linear transformation between the two domains
by minimizing  the distance between the empirical expectations
of source and target data  distributions integrated within a kernel embedding. In addition,  to  put
less emphasis  on the source instances  that are irrelevant
to classify the target data,   instance re-weighing is employed.

The feature transformation proposed by in \cite{ChenICML12Marginalized}
exploits the correlation between the source and target set to  learn a robust  representation 
by  reconstructing the original features from their noised counterparts. The method, called
Marginalized Denoising Autoencoder (MDA),  
is based on a quadratic loss and a drop-out noise level that factorizes over all feature dimensions. 
This  allows the method to avoid explicit data corruption
by marginalizing out the noise and to have a closed-form solution for 
the feature transformation. Note that it is straightforward to stack together several layers with optional non-linearities  
between layers  to obtain a  multi-layer network  with the parameters for each layer obtained in a 
single forward pass (see Algorithm \ref{Limited_alg_mda}). \\

\begin{algorithm}[t]
\caption{\small Stacked Marginalized Denoising Autoencoder (sMDA)~\cite{ChenICML12Marginalized}.}
\label{Limited_alg_mda}
\begin{algorithmic}[1]
\Input{Source data $\bfX^s$, target data $\bfX^t$}
\Input{Parameters:  $p$ (noise level), $\omega$ (regularizer) and $k$ (number of stacked layers)}
\State{$\bfX= [\bfX^s, \bfX^t]$, \,\, $\bfS = \bfX^\top \bfX$, \,\, and \,\, $\bfX_0=\bfX$;}
\State{$\bfP= (1-p) \bfS$  \,\, and  \,\, $\bfQ=(1-p)^2 \bfS +p(1-p)diag(\bfS)$}
\State{$\bfW =(\bfQ+ \omega \eye_{D})^{-1} \bfP$.}
\State{(Optionally), stack $K$ layers with $\bfX_{(k)}=\tanh( \bfX_{(k-1)} \bfW^{(k)})$.}
\Output{Denoised features $\bfX_k$.}
\end{algorithmic}
\end{algorithm}

In general, the above mentioned  methods  learn the transformation without using any class label. 
After projecting the data in the new space, any classifier trained on the source set 
can be used to predict labels for the target data. The model often works even better if in addition a small set of the target 
examples are hand-labeled (SS adaptation).  
The class labels can also be used to learn a better transformation.  Such methods, called 
supervised feature transformation based DA methods,  to learn the transformation exploit class labels, 
either only from the source or also from the target (when available). 
When only the source class labels are exploited, 
the method  can still be applied  to the US scenario, while methods using also target labels are 
designed for the SS case.   \\

\noindent {\bf Supervised feature transformation.}  
Several  unsupervised feature transformation  methods, cited above,  have been extended 
to capitalize on class labels to learn a better
transformation. Among these extensions,  we can mention the
Semi-Supervised TCA \cite{PanTNN11Domain,MatasciTGRS15Semisupervised}  where 
the objective function that is 
minimized contains a label dependency term in addition to  the distance between the domains and the 
manifold regularization term. The  label dependency term  has the role of maximizing the alignment of the projections
with the  source labels and, when available,  target labels.

 Similarly, in \cite{CsurkaTASKCV16Unsupervised} a quadratic regularization term,  relying on the 
 pretrained source classifier,  is added into the MDA framework \cite{ChenICML12Marginalized}, in order to  keep
 the denoised source data  well classified. 
Moreover, the domain denoising and  cross-domain classifier can be learned jointly
 by iteratively solving  a Sylvester linear system to estimate the transformation and 
 a linear system to get the classifier in 
 closed form\footnote{Code at \url{https://github.com/sclincha/xrce_msda_da_regularization}}.

  To take advantage of  class labels,  the distance between each source sample and
its corresponding class means is added  as regularizer into the DIP \cite{BaktashmotlaghICCV13Unsupervised}
respectively SIE model \cite{BaktashmotlaghCVPR14Domain}. This term  
encourages the  source samples from  the same class to be clustered in the latent space.  
  The Adaptation Regularization based Transfer 
  Learning\footnote{Code at \url{http://ise.thss.tsinghua.edu.cn/~mlong/doc/adaptation-regularization-tkde14.zip}} \cite{LongTKDE14Adaptation}   
  performs  DA  by optimizing simultaneously  the structural risk functional, 
 the joint distribution matching between domains and the manifold consistency.
 The Max-Margin Domain Transform\footnote{Code at \url{https://cs.stanford.edu/~jhoffman/code/Hoffman_ICLR13_MMDT_v3.zip}}
 \cite{HoffmanICLR13Efficient} 
optimizes both the transformation and classifier parameters jointly, by introducing
an efficient cost function based on the misclassification loss.

\begin{figure}[ttt]
\begin{center}
\includegraphics[width=.8\textwidth]{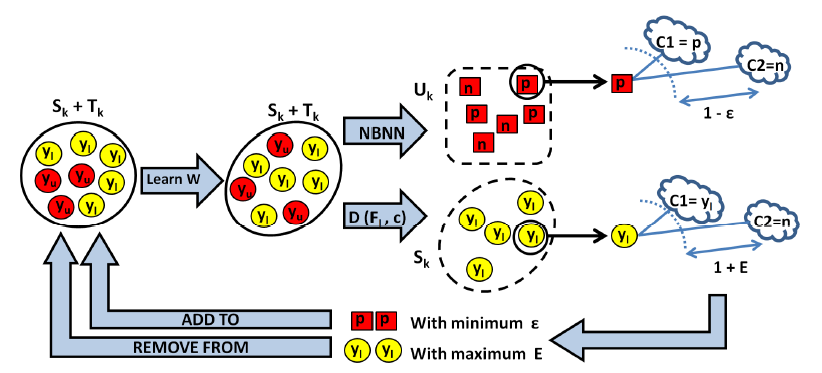}
\caption{The NBNN-DA adjusts the image-to-class distances by tuning the per class metrics and 
iteratively making the metric progressively more suitable for the target.
(Image: Courtesy to T. Tommasi~\cite{GopalanICCV11Domain})}
\label{fig.NBNNDA}
\end{center}
\end{figure}

Another set of methods extend marginal distribution discrepancy minimization   to 
conditional distribution involving  data labels from  the source and  class predictions from the target.  
Thus,  \cite{ZhongKDD09Cross}  proposes an adaptive kernel approach that maps the marginal 
distribution of the target and source sets into a common kernel space, 
and use a sample selection strategy to draw conditional probabilities between the two domains closer. 
The Joint Distribution 
Adaptation\footnote{Code at \url{http://ise.thss.tsinghua.edu.cn/~mlong/doc/joint-distribution-adaptation-iccv13.zip}}
 \cite{LongICCV13Transfer} jointly adapts  the marginal distribution
through a principled (PCA based) dimensionality reduction procedure 
and the conditional distribution between the domains.  \\

\noindent {\bf Metric learning based  feature transformation.} These methods are particular 
supervised feature transformation
methods that involves that at least a limited set of target labels are available,  and they use 
metric learning techniques  to bridge the relatedness
between the source and target domains.  Thus, \cite{ZhaIJCAI09Robust} proposes  distance metric learning
with either log-determinant  or manifold regularization to adapt face recognition models between subjects. 
\cite{SaenkoECCV10Adapting} uses the Information-Theoretic 
Metric Learning from \cite{DavisICML07Informationtheoretic}  to  define  a common  distance metric across different domains.
This method was further extended in \cite{KulisCVPR11What} by incorporating non-linear kernels, which  enable
the model to be  applicable  to the heterogeneous case (\ie different source and target 
representations).

The metric learning for  Domain Specific Class Means (DSCM)
\cite{CsurkaTASKCV14Domain}  learns a transformation of the feature space which, for each 
instance minimizes the  weighted  soft-max distances to the corresponding domain specific class means.
 This allows in the 
projected space to decrease  the intraclass  and to increase the interclass distances  
(see also Figure \ref{fig.DSCMML}). 
This was extended with an active learning component by the
 Self-adaptive Metric Learning Domain Adaptation (SaML-DA) \cite{CsurkaTASKCV14Domain} framework, where 
 the target training set is iteratively increased  with  labels predicted with DSCM and 
 used to refine the current metric.  SaML-DA  was inspired by the Naive  
Bayes Nearest Neighbor based Domain Adaptation\footnote{Code at \url{http://www.tatianatommasi.com/2013/DANBNNdemo.tar.gz}}
 (NBNN-DA) \cite{TommasiICCV13Frustratingly} framework, 
which  combines metric learning and NBNN classifier  to adjust the 
instance-to-class distances by  progressively making the metric  more suitable for the target domain
(see Figure \ref{fig.NBNNDA}). The main idea behind both methods, SaML-DA and NBNN-DA, is to replace
at each iteration the  most  ambiguous source example of each class
by the target example  for which the classifier (DSCM respectively NNBA) is the most confident 
for the given class.  \\

\begin{figure}[ttt]
\begin{center}
\includegraphics[width=1\textwidth]{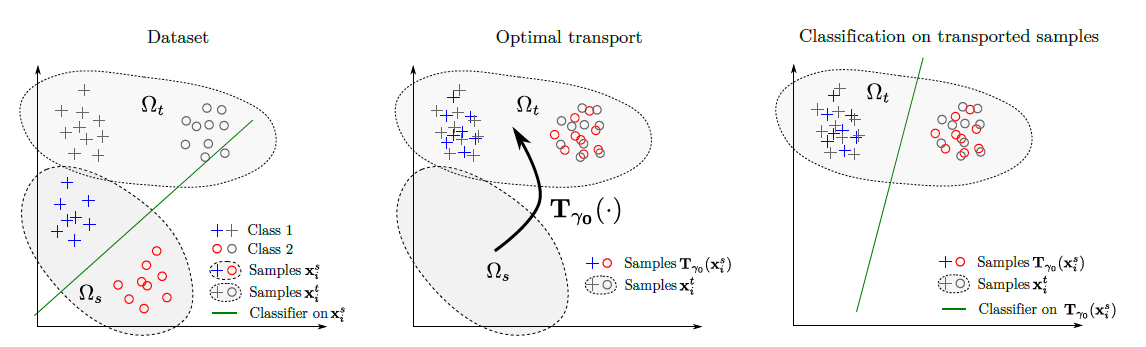}
\caption{The OTDA~\cite{CourtyX15Optimal} consider a  local transportation plan
for each sample in the source domain to transport the training
samples close to the target examples. (Image: Courtesy to N. Courty.)}
\label{fig.OTDA}
\end{center}
\end{figure}

\noindent {\bf Local feature transformation.}  The previous methods 
 learn a global  transformation to be applied to each source and target example. 
In contrast, the 
Adaptive Transductive Transfer Machines (ATTM)~\cite{FarajidavarBMVC14Adaptive}
complements the global transformation with a  sample-based transformation to refine the 
probability density function of the source instances assuming that the transformation from the source to the
target domain is locally linear. This is achieved by representing the target set by a Gaussian Mixture Model and 
learning an optimal translation parameter  that maximizes
the likelihood of the translated source  as a posterior.

Similarly,   the Optimal Transport for Domain Adaptation \cite{CourtyX15Optimal}, 
 considers a local transportation plan for each source example. 
The model  can be seen as a graph matching problem, where the final 
coordinates of each sample  are found
by mapping the source samples to the target ones, whilst respecting the marginal distribution of the target domain
(see Figure~\ref{fig.OTDA}).
To exploit class labels, a regularization term with group-lasso is added 
 inducing, on one hand, group sparsity and, on  another hand, constraining source samples 
 of the same class to remain close
during the transport. \\

\begin{figure}[ttt]
\begin{center}
\includegraphics[width=1\textwidth]{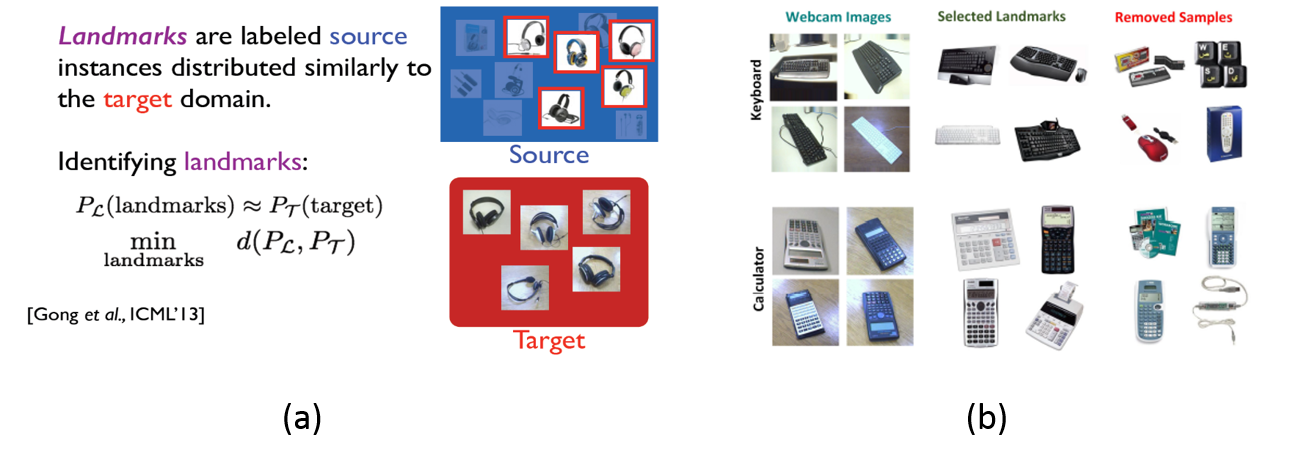}
\caption{Landmarks selected for the task {\em amazon versus webcam} 
using the popular  Office31 dataset ~\cite{SaenkoECCV10Adapting}  with 
(a) MMD~\cite{GongICML13Connecting}  
and  (b) the Hellinger distance on the statistical manifold~\cite{BaktashmotlaghCVPR14Domain}.}
\label{fig.Landmark}
\end{center}
\end{figure}

\noindent{\bf Landmark selection.} In order to improve the feature learning process, 
several methods have been proposed with the aim of selecting  the most relevant instances 
from  the source,  so-called landmark examples,  to be used to train the  adaptation model
(see examples   in Figure \ref{fig.Landmark}).
Thus, \cite{GongICML13Connecting}  proposes to minimize a variant of the MMD
to identify good  landmarks by creating a set of auxiliary tasks that offer multiple views of the original
problem\footnote{Code at \url{http://www-scf.usc.edu/~boqinggo/domain_adaptation/landmark_v1.zip}}. 
The Statistically Invariant Sample Selection \cite{BaktashmotlaghCVPR14Domain}, 
uses the Hellinger distance on the statistical manifold instead of  MMD. The selection is forced to 
keep the proportions of the source samples per class the
same as in the original data. Contrariwise to these approaches, the Multi-scale Landmark 
Selection\footnote{Code at \url{http://home.heeere.com/data/cvpr-2015/LSSA.zip}}
\cite{AljundiCVPR15Landmarks}  does not require any class labels. It 
takes each instance independently and considers it as being a good candidate  if the Gaussian 
distributions of the source examples and of the target points centered on the instance 
are similar over a set of different scales (Gaussian variances). 

Note that the landmark selection process, although strongly related to instance re-weighting methods with binary weights, 
can be rather seen as data preprocessing and hence complementary to the adaptation process. \\

\subsection{Multi-source domain adaptation} 
\label{sec.multisource}

Most of the above mentioned methods were designed for a single source \vs target case. When 
multiple sources are available, they can be concatenated to form  a {\em single} source set,  but 
because the possible shift between the different source domains, this
 might not be  always a good option. Alternatively, the models built for each {\em source-target} 
pair (or their results)  can be combined to make a final decision. 
However, a better option might be to build multi-source DA models which, relying only on the {\em a priori known} domain labels,
  are able to exploit the specificity of each source domain. 

Such methods are  the Feature Augmentation (FA) \cite{DaumeX09Frustratingly} 
and the A-SVM \cite{YangMM07Crossdomain},  already mentioned  in  Section \ref{sec.homogeneous},
 both exploiting naturally the multi-source aspect of the dataset. Indeed in 
 the case of FA,  extra feature sets,  one for each source domain, concatenated to the representations, 
 allow to learn source specific properties  shared  between a given source and the target. 
 The A-SVM  uses an ensemble of source specific auxiliary classifiers  to adjust the parameters of the target  classifier.

\begin{figure}[ttt]
\begin{center}
\includegraphics[width=0.9\textwidth]{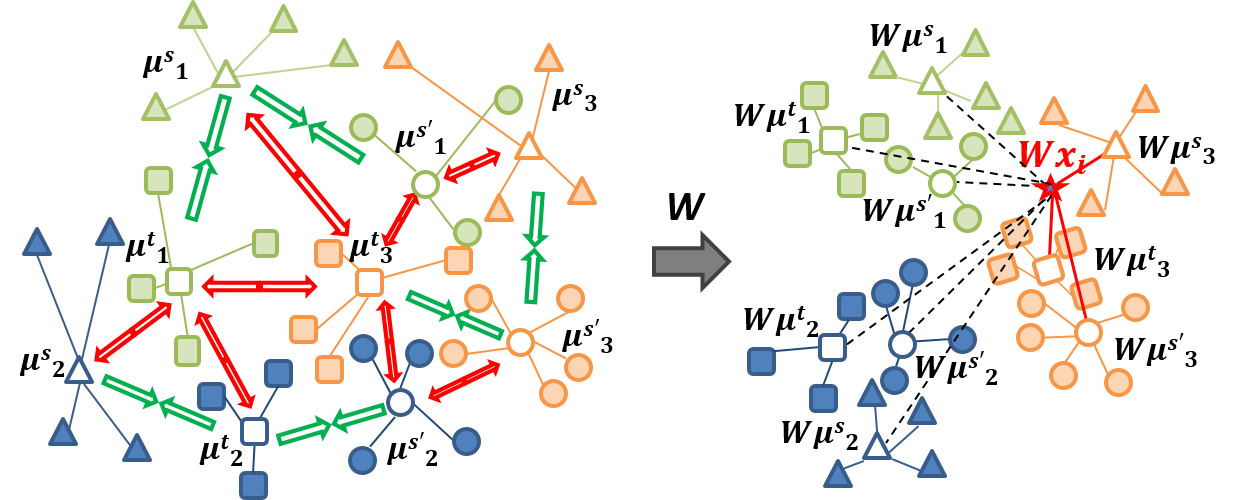}
\caption{Metric learning for the DSCM classifier, where $\mu^s_{c_i}$ and $\mu^{s'}_{c_i}$  represent 
source specific class means and $\mu^t_{c_i}$ class means in the target domain. The feature transformation $\bfW$ is learned by 
minimizing for each sample the  weighted  soft-max distances to the corresponding domain specific class means in the projected space.} 
\label{fig.DSCMML}
\end{center}
\end{figure}

 Similarly, the Domain Adaptation Machine \cite{DuanTNNLS12Domain}  l
 leverages a set of source  classifiers  by  the integration of  domain-dependent regularizer
 term which is  based on a smoothness assumption.  The model forces the target classifier to share similar 
 decision values with the relevant source classifiers on the unlabeled target  instances.
The Conditional Probability based Multi-source Domain Adaptation (CP-MDA)
approach \cite{ChattopadhyaySIGKDD11Multisource} extends 
the above idea  by  adding weight values for each source classifier based on conditional distributions. 
The DSCM proposed  in \cite{CsurkaTASKCV14Domain}   relies on domain specific class means both to
learn the metric but also to predict the target class labels (see  illustration in Figure \ref{fig.DSCMML}). 
The domain regularization and classifier based regularization terms  of the  
extended MDA \cite{CsurkaTASKCV16Unsupervised} are both sums of source specific components.

\begin{figure}[ttt]
\begin{center}
\includegraphics[width=.75\textwidth]{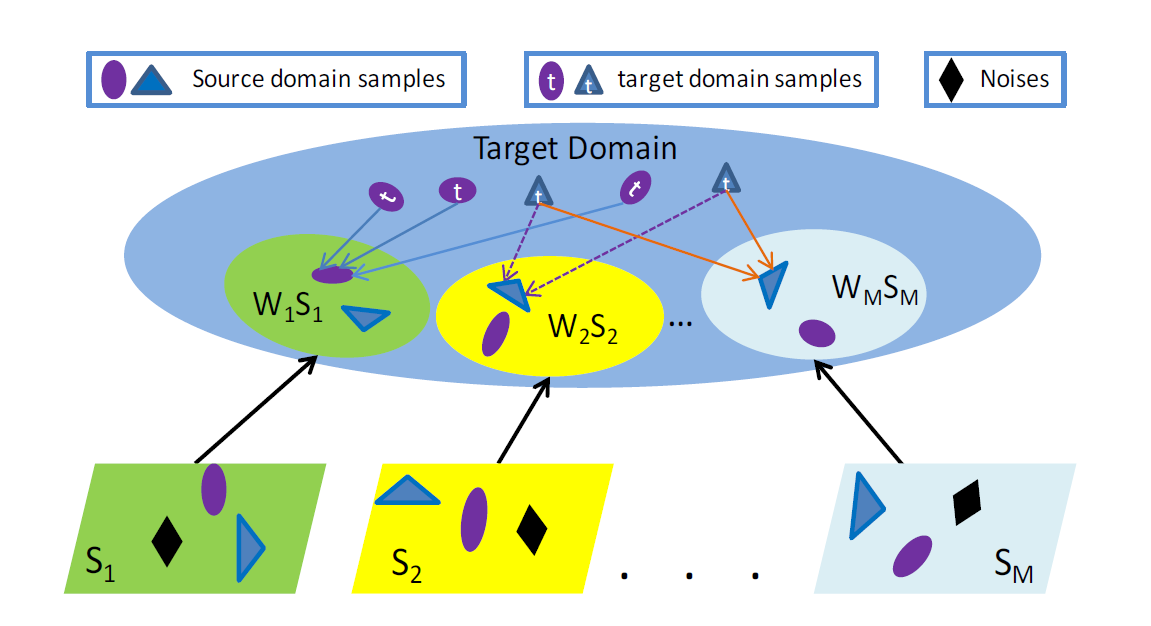}
\caption{The RDALRR~\cite{JhuoCVPR12Robust}   transforms each source domain  into an intermediate
representation such that the transformed samples
can be linearly reconstructed from the target samples.  (Image: Courtesy to I.H. Jhuo.)}
\label{fig.RDALRR}
\end{center}
\end{figure}

The  Robust DA via Low-Rank Reconstruction (RDALRR) \cite{JhuoCVPR12Robust} transforms 
each source domain  into an intermediate
representation such that the transformed samples
can be linearly reconstructed from the target ones. Within each
source domain, the intrinsic relatedness of the reconstructed samples is imposed by using  a 
low-rank structure where  the 
 outliers are identified  using  sparsity constraints. By enforcing different source domains 
to have jointly low ranks,  a compact source sample set is formed with a
distribution  close to the target domain (see Figure \ref{fig.RDALRR}).

To better take advantage of  having multiple source domains,  extensions to methods   
previously  designed for a single source \vs target case 
were proposed  in \cite{GopalanPAMI14Unsupervised,CaseiroCVPR15Beyond,HoffmanECCV12Discovering,YaoCVPR10Boosting}. 
Thus, \cite{GopalanPAMI14Unsupervised}  describes  a multi-source version of the GFS \cite{GopalanICCV11Domain}, 
which was further extended in \cite{CaseiroCVPR15Beyond} to  
the Subspaces by Sampling Spline Flow   approach. The latter  uses smooth polynomial functions determined
by splines on the manifold to interpolate between different source and the target domain. 
\cite{HoffmanECCV12Discovering} combines\footnote{Code at \url{https://cs.stanford.edu/~jhoffman/code/hoffman_latent_domains_release_v2.zip}} 
constrained clustering algorithm,  
used to identify automatically source domains in a large data set, with a multi-source extension 
of the Asymmetric Kernel Transform \cite{KulisCVPR11What}. 
\cite{YaoCVPR10Boosting} efficiently extends the TrAdaBoost \cite{DaiICML07Boosting}  
to multiple source domains. \\

\noindent{\bf Source domain weighting.}
When multiple sources are available,  it is desired to select those domains that provide the best information transfer 
and to remove the ones that have more likely negatively impact on the final model. 
Thus,  to  down-weight the effect of  less related source domains, in \cite{GeSDM13LHandling} 
first the available labels are   propagated  within clusters obtained by  spectral clustering
 and then to  each source cluster  a Supervised Local Weight (SLW) is assigned 
 based on the   percentage of label matches between predictions made by a source model 
 and those made by label propagation.

In the Locally Weighted Ensemble framework \cite{GeSDM13LHandling}, the model weights
are  computed  as a  similarity between the local neighborhood graphs   centered on  source  and target instances. 
The CP-MDA \cite{ChattopadhyaySIGKDD11Multisource}, mentioned above, 
 uses a weighted combination of source learners, where the weights are estimated as a function of conditional probability
differences between the source and target domains. The Rank of Domain  value 
defined in \cite{GongCVPR12Geodesic}   measures the relatedness between each source and 
target domain as the  KL divergences between data distributions  once the data is projected into the latent subspace. 
The Multi-Model Knowledge Transfer \cite{TommasiCVPR10Safety} minimizes the  negative transfer by giving  higher weights
to the most related linear SVM source  classifiers. These weights are determined through a leave one out learning process.

\subsection{Heterogeneous domain adaptation}
\label{sec.heterogeneous}

{\em Heterogeneous   transfer learning} (HTL) refers to the setting where the
representation spaces  are different for the source and target domains 
($\calX^t \neq \calX^s$ as defined in  Section \ref{sec.definitions}).  
As a particular case,  when   the tasks are assumed to be the same,  \ie $\calY^s=\calY^t$, 
we refer to it as {\em heterogeneous domain adaptation} (HDA). 

Both HDA and HTL are strongly related to  multi-view learning \cite{XuX13Survey,WangICML15Deep}, where 
the presence of multiple information sources gives an opportunity to learn better representations (features)
 by analyzing the views simultaneously. This makes possible to 
solve the  task when not all the views are available. 
Such situations appear  when  processing  simultaneously audio and video \cite{ChaudhuriICML09Multiview},  
 documents containing both image and text 
(\eg web pages or photos with  tags or comments) \cite{HardoonNC05Canonical,SocherCVPR10Connecting,YanCVPR15Deep},  
images acquired with depth information \cite{HoffmanCVPR16Learning}, \etc.
We can also have multi-view settings when the views have the same modalities (textual, visual, audio),
such as  in the case of  parallel text corpora in different languages \cite{VinokourovNIPS03Using,FaruquiEACL14Improving},
photos of the same person taken across different poses, illuminations and
expressions \cite{YangICCV11Fisher,SharmaCVPR11Bypassing,SharmaCVPR12Generalized,QiuECCV12Domain}. 

\begin{figure}[ttt]
\begin{center}
\includegraphics[width=.6\textwidth]{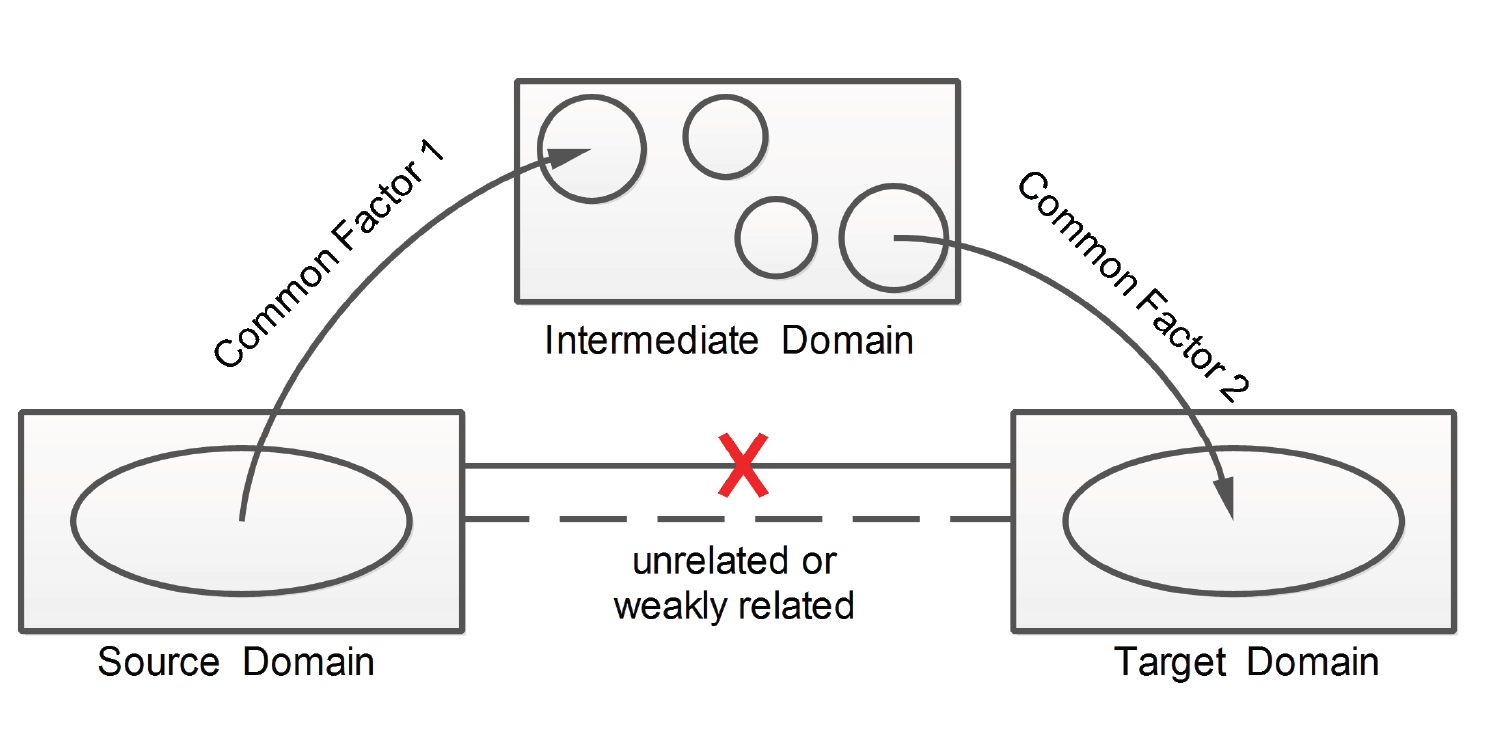}
\caption{Heterogeneous DA  through an intermediate domain allowing to bridge the gap between  
features representing the two domains.  For example, when the source domain contains text and the target  images, 
the intermediate domain can be built from a set of crawled Web pages containing both text and images. 
(Image courtesy B. Tan~\cite{TanSIGKDD15Transitive}).} 
\label{fig.cooccurence}
\end{center}
\end{figure}

Multi-view learning  assumes that at training  time for the same data instance multiple views
from complementary information sources are available (\eg a person is identified by photograph, fingerprint,
signature or iris). Instead, in the case of  HTL and HDA, the challenge comes from the fact that 
we have one view at training and another one at test time.  Therefore, one set of methods  
proposed to solve HDA relies on some multi-view auxiliary data\footnote{When the bridge is to be done 
between visual and textual representations, a common practice is to crawl the Web for 
pages containing both text and images in order to build such intermediate multi-view data.}    
to bridge the gap between the domains  (see Figure \ref{fig.cooccurence}).     \\

\noindent{\bf Methods relying on auxiliary domains.} 
These methods principally exploit feature co-occurrences (\eg between words and visual features) in the  
multi-view auxiliary domain. As such, the  Transitive Transfer Learning \cite{TanSIGKDD15Transitive}  selects 
an appropriate domain from a large data set  guided by domain complexity
and,  the distribution differences between the original domains (source and target) and the selected one (auxiliary). 
Then,  using Non-negative Matrix Tri-factorization \cite{DingKDD05Orthogonal}, 
feature clustering and label propagation is performed simultaneously through the intermediate domain. 

The  Mixed-Transfer approach \cite{TanSICDM14MixedTransfer} builds a 
joint transition probability graph of mixed instances and features, considering the data in
the source,  target and  intermediate domains.  The label propagation on the graph is done  
by a random walk process to overcome  the 
data sparsity.  In \cite{ZhuAAAI11Heterogeneous}
the representations of the target images are enriched
 with semantic concepts extracted from  the intermediate 
 data\footnote{Code available at \url{http://www.cse.ust.hk/\%7Eyinz/htl4ic.zip}} 
through a  Collective Matrix Factorization \cite{SinghKDD08Relational}.

\cite{QiWWW11Towards} proposes to build a translator 
function\footnote{Code available at \url{http://www.ifp.illinois.edu/\%7Eqi4/TTI_release_v1.zip}}
between the source and target domain by learning directly  the product of the two transformation matrices 
that map each domain into a common (hypothetical) latent topic built on the co-occurrence data. 
Following the principle of parsimony, they encode as few topics as possible in order to be able to match 
text and images. The semantic labels are propagated from the labeled text corpus to 
unlabeled new images by a cross-domain label propagation mechanism using the built translator.
In \cite{YangTNNLS15Learning}  the co-occurrence data is  represented by the principal components
computed  in each feature space  and  a Markov Chain Monte Carlo \cite{AndrieuML03Introduction}
is employed to construct a directed cyclic network 
where each node is a domain and each edge weight represents the conditional
 dependence between the corresponding  domains defined by the transfer weights. 
 
\begin{figure}[ttt]
\begin{center}
\includegraphics[width=1.1\textwidth]{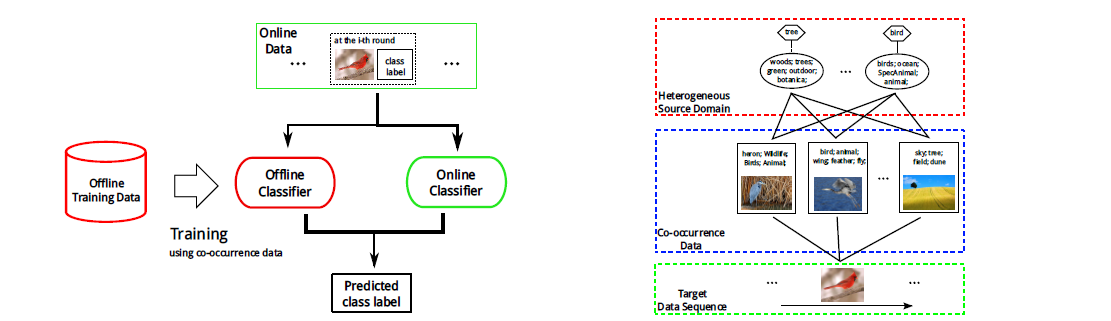}
\caption{Combining the online classifier with the offline classifier (right) and transfer the knowledge 
through  co-occurrences  data in the heterogeneous intermediate domain (left). 
 (Image: Courtesy to Y. Yan~\cite{YanTASKCV16Online})}
\label{fig.OnlineHDA}
\end{center}
\end{figure}

 \cite{YanTASKCV16Online} studies online HDA, where offline labeled data from a source domain is
transferred to enhance the online classification performance for the  target domain. The main idea is to build an offline
classifier based on heterogeneous similarity  using labeled data from a source domain and unlabeled co-occurrence data 
 collected from Web pages and social networks (see Figure \ref{fig.OnlineHDA}). 
 The  online target classifier  is combined with  the
offline source classifier using Hedge weighting strategy, used in Adaboost \cite{FreundJCSS97Decision}, 
to update their weights for ensemble prediction. \\

Instead of relying on external data to bridge the data representation gap, several  HDA methods 
exploit directly the data distribution in the source and target domains willing to remove
simultaneously  the gap between  the feature representations and minimizing the data distribution shift. This is done 
  by  learning  either a projection for each domain into a domain-invariant common latent space, referred to as   
{\em symmetric transformation} based HDA\footnote{These methods can  be  used even if the source and target data
are represented in the same feature space, \ie  $\calX^t =\calX^s$.
Therefore, it is not surprising that several methods are direct extensions of  homogeneous DA methods 
described in Section \ref{sec.homogeneous}.},  or a transformation from the source space towards the target space, 
called {\em asymmetric transformation} based HDA.  These approaches require at least a limited amount of labeled target examples 
(semi-supervised DA). \\

\noindent{\bf Symmetric feature transformation.}
The aim of symmetric transformation based HDA approaches  is to learn projections  for both the source and  target spaces 
into a common latent (embedding) feature space  better suited to learn the task for the target.
These methods are  related,  on one hand, to the
feature transformation based  homogeneous  DA methods described in Section \ref{sec.homogeneous} 
and,  on another hand,  to  
multi-view embedding \cite{HardoonNC05Canonical,NgiamICML11Multimodal,SharmaCVPR12Generalized,GongIJCV14multiview,CaoX16Generalized,WangCVPR16Learning},
 where  different views are embedded in a common latent space.  
 Therefore, several DA methods  originally designed for the homogeneous case, 
 have been inspired by the multi-view embedding approaches
and  extended to heterogeneous data.

\begin{figure}[ttt]
\begin{center}
\includegraphics[width=0.9\textwidth]{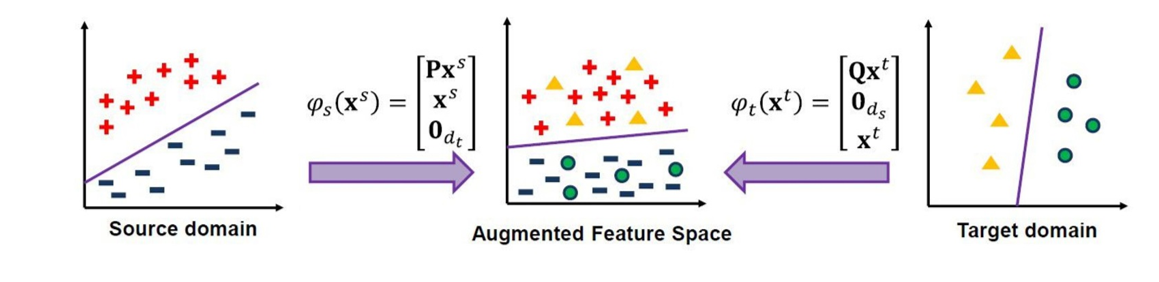}
\caption{The HFA~\cite{DuanPAMI12Learning} is seeking for an optimal common space 
while simultaneously learning a discriminative SVM
classifier. (Image: Courtesy to Dong Xu.)}
\label{fig.HFA}
\end{center}
\end{figure}

  As such,    the Heterogeneous Feature 
  Augmentation\footnote{Code available at \url{https://sites.google.com/site/xyzliwen/publications/HFA_release_0315.rar}} 
  (HFA)~\cite{DuanPAMI12Learning}, 
  prior to data augmentation, embeds the source and  target  into  a common latent space 
(see  Figure~\ref{fig.HFA}).  In order to avoid the explicit projections, the
transformation metrics  are computed by  the minimization of the structural risk functional  
of SVM expressed as a function of these projection  matrices.  The
final target prediction function is  computed by an alternating optimization algorithm to
simultaneously solve the dual SVM and to find the optimal transformations. 
This model was further extended  in \cite{LiPAMI14Learning},  where each projection matrix  
is decomposed into a linear combination of a set of 
rank-one positive semi-definite matrices and they  are combined within a  Multiple Kernel Learning approach.

The Heterogeneous Spectral Mapping \cite{ShiICDM10Transfer}  unifies different feature spaces using spectral 
embedding where the similarity between the  domains in the latent space is  maximized
with the constraint to preserve the original structure of the data.  Combined with a  source 
sample selection strategy,  a Bayesian-based approach is applied 
to model the relationship between the different output spaces.

\cite{XiaoECML15Semisupervised}  present a semi-supervised subspace co-projection method, which
addresses heterogeneous multi-class DA. It is based on discriminative subspace learning  
and exploit  unlabeled data to enforce an MMD criterion across domains in the projected subspace. 
They  use  Error Correcting Output Codes  (ECOC) to address the multi-class aspect
and to enhance the discriminative informativeness of the
projected subspace.  The Semi-supervised Domain Adaptation with Subspace Learning  \cite{YaoCVPR15Semisupervised}  
jointly explores invariant low-dimensional structures across domains to correct data 
distribution mismatch and leverages available 
unlabeled target examples to exploit the underlying intrinsic information in the 
target domain.

\begin{figure}[ttt]
\begin{center}
\includegraphics[width=.75\textwidth]{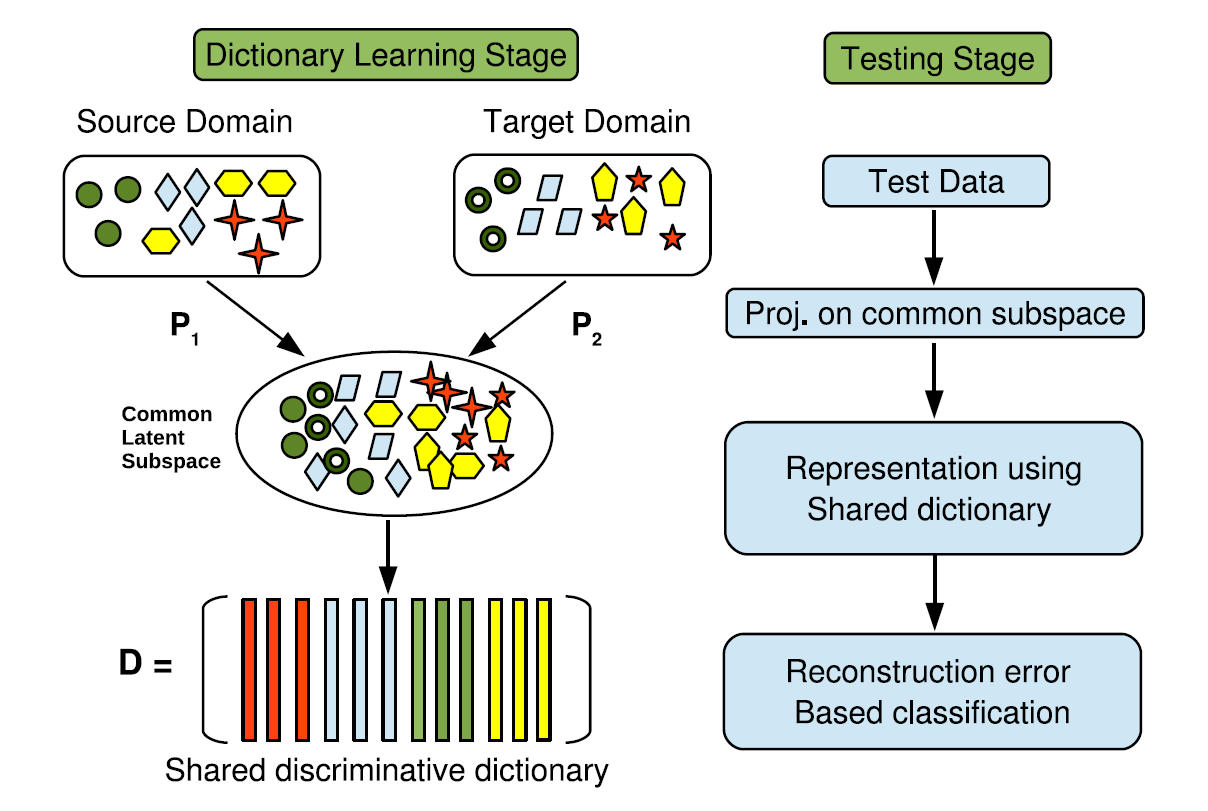}
\caption{The SDDL proposes to learn a  dictionary in a latent common subspace 
while maintaining the manifold structure of the data.
 (Image: Courtesy to S. Shekhar~\cite{ShekharCVPR13Generalized})}
\label{fig.SDDL}
\end{center}
\end{figure}

To deal with both  domain shift  and heterogeneous data,  
the Shared Domain-adapted Dictionary 
Learning\footnote{Code available at \url{http://www.umiacs.umd.edu/~pvishalm/Codes/DomainAdaptDict.zip}}  
  (SDDL) \cite{ShekharCVPR13Generalized} 
 learns a class-wise discriminative dictionary in the latent  projected space  (see  Figure \ref{fig.SDDL}).
 This is done by jointly learning the dictionary and the projections
of the data from both domains onto a common low-dimensional space,
while maintaining the manifold structure of data represented  by sparse
linear combinations of dictionary atoms. 

The Domain Adaptation Manifold Alignment (DAMA) \cite{WangIJCAI11Heterogeneous}   models 
each domain  as a manifold and  creates a separate mapping function to transform the heterogeneous 
input space into a common latent  space while preserving the underlying structure of each domain. 
This is done by representing each domains with a Laplacian that captures the closeness of 
the instances sharing the same label. 
 The RDALRR \cite{JhuoCVPR12Robust}, mentioned above (see also Figure~\ref{fig.RDALRR}),  
 transforms each source domain  into an intermediate
representation such that the source samples linearly reconstructed   from the target samples  are 
enforced to be related to each other under a low-rank structure.
Note that  both DAMA and RDALRR are multi-source HDA approaches.  \\

\noindent {\bf Asymmetric feature transformation.} In contrast to symmetric transformation based HDA, 
these methods aim to learn a projection of the source features into the target space such that  
the distribution mismatch within each class is minimized. Such method is the  
Asymmetric Regularized Cross-domain 
Transformation\footnote{Code available at \url{http://vision.cs.uml.edu/code/DomainTransformsECCV10_v1.tar.gz}} 
\cite{KulisCVPR11What} that
utilizes an objective function responsible for  the domain invariant transformation  
learned in a non-linear Gaussian RBF kernel space.
The Multiple Outlook MAPping algorithm \cite{HarelICML11Learning} 
 finds the transformation matrix by  singular value decomposition process 
 that encourage the marginal distributions within the
classes to be aligned while maintaining the structure of the data. It requires a
 limited amount of labeled target data for each class to be 
 paired with the corresponding source classes. 

\cite{ZhouAISTATS14Heterogeneous} proposes a sparse and class-invariant feature mapping that 
 leverages the weight vectors of the binary classifiers learned
in the source and target domains. This is done by  considering the learning task 
as a Compressed Sensing \cite{DonohoTIT06Compressed} problem and using the  ECOC scheme to generate 
a sufficient number of binary classifiers given the set of classes.

\begin{figure}[ttt]
\begin{center}
\includegraphics[width=0.85\textwidth]{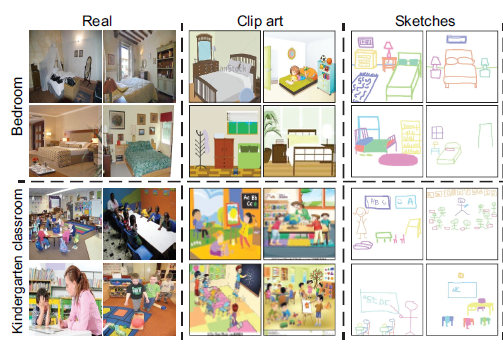}
\caption{Examples from the Cross-Modal Places Dataset (CMPlaces)
dataset proposed in \cite{CastrejonCVPR16Learning}. (Image: Courtesy to L. Castrej\'{o}n.)}
\label{fig.PlacesDA}
\end{center}
\end{figure}

\section{Deep domain adaptation methods}
\label{sec.deep}

With the recent progress in image categorization due to 
deep convolutional architectures - trained in a fully supervised fashion on large scale annotated 
datasets, in particular on part of  ImageNet~\cite{RussakovskyIJCV15Imagenet} -
allowed a significant improvement of the categorization accuracy over previous state-of-the art solutions. 
Furthermore, it was shown that features extracted from
the activation layers of these deep convolutional networks can be re-purposed
to novel tasks~\cite{DonahueICML14Decaf} even when the new tasks   differ significantly from the task originally
used to train the model.

Concerning domain adaptation, baseline methods without adaptation obtained  using  
features generated by deep models~\footnote{Activation layers 
extracted from  popular CNN models, such as  AlexNet~\cite{KrizhevskyNIPS12Imagenet},  VGGNET~\cite{SimonyanX14Very},  
ResNet~\cite{HeX15Deep}  or GoogleNet~\cite{SzegedyCVPR15Going}.}  
on the two most popular benchmark datasets  Office (OFF31)~\cite{SaenkoECCV10Adapting}  
and Office+Caltech (OC10)~\cite{GongCVPR12Geodesic} 
outperform by a large margin the shallow DA methods using
the SURFBOV features originally provided with these datasets. Indeed, the 
results obtained with  such Deep Convolutional Activation 
Features\footnote{Code to extract features available at \url{https://github.com/UCBAIR/decaf-releas}}
(DeCAF)~\cite{DonahueICML14Decaf} 
even without any adaptation  to the target are significantly better  that the results obtained with any 
DA method based on SURFBOV~\cite{ChopraWREPL13Dlid,DonahueICML14Decaf,SunAAAI16Return,CsurkaTASKCV16Unsupervised}.
As shown also in~\cite{BengioPAMI13Representation,YosinskiNIPS14How}, this suggests
that deep neural networks  learn more abstract and robust representations, encode category level information and remove, 
to a certain measure,  the domain bias~\cite{DonahueICML14Decaf,SunAAAI16Return,CsurkaTASKCV16Unsupervised,SaxenaTASKCV16Heterogeneous}. 

Note  however that in OFF31 and OC10 datasets the images  remain relatively similar to the images 
 used to train these models  (usually datasets
from the ImageNet Large-Scale Visual Recognition Challenge \cite{RussakovskyIJCV15Imagenet}). 
In contrast, if we consider category models between \eg images and paintings, drawings, clip art or 
sketches (see see  examples from the CMPlaces 
dataset\footnote{Dataset available at \url{http://projects.csail.mit.edu/cmplaces/}}   
in Figure~\ref{fig.PlacesDA}), the models have more difficulties to handle the domain
differences \cite{KlareICB12STowards,CrowleyBMVC14State,CrowleyCVAA14Search,CastrejonCVPR16Learning} 
and alternative solutions are necessary.

Solutions proposed in the literature to exploit deep models can be grouped into three main categories. 
The first group considers the CNN models to extract vectorial features to be used by 
the shallow DA methods. The second solution is to train or fine-tune the deep network
on the source domain, adjust it to the new task, and use the model to predict class labels for target instances. 
Finally, the most promising methods are based on deep
learning architectures designed for DA.

\begin{figure}[ttt]
\begin{center}
\includegraphics[width=1\textwidth]{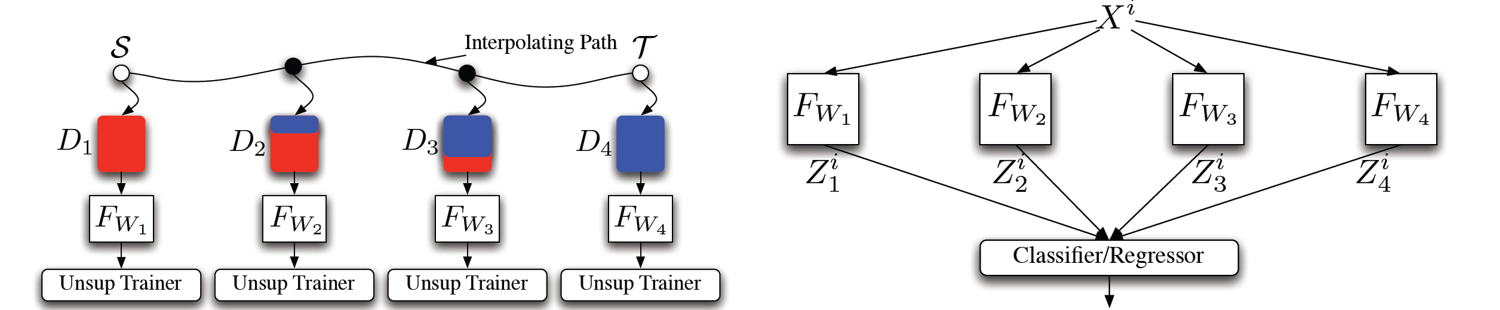}
\caption{The DLID model aims in interpolating between domains based 
on the amount of source and target data used to train each  model. 
 (Image courtesy S. Chopra~\cite{ChopraWREPL13Dlid}).} 
\label{fig.DLID}
\end{center}
\end{figure}

\noindent {\bf Shallow methods with  deep features.}
The first, naive solution is to consider the deep network as feature extractor, where 
the activations of a layer or several layers of the deep architecture is considered as 
representation for the input image. 
These  Deep Convolutional Activation Features (DeCAF) \cite{DonahueICML14Decaf}  
extracted from both source and target examples 
can then be used within any  shallow DA method  described in Section \ref{sec.shallow}. 
For example,  Feature Augmentation \cite{DaumeX09Frustratingly}, 
Max-Margin Domain Transforms \cite{HoffmanICLR13Efficient}  and  Geodesic Flow Kernel \cite{GongCVPR12Geodesic} 
were applied to DECAF features in \cite{DonahueICML14Decaf}, 
Subspace  Alignment \cite{FernandoICCV13Unsupervised} and Correlation Alignment in \cite{SunAAAI16Return}. 
\cite{CsurkaTASKCV16Unsupervised}  experiments with DeCAF features within the extended 
MDA framework, while \cite{SaxenaTASKCV16Heterogeneous} explores various metric learning 
approaches to align deep features extracted from RGB face images (source) and NIR or sketches (target). 

In general, these DA methods allow to further improve the classification accuracy compared to
 the baseline  classifiers trained only on the source data  with  these DeCAF 
features \cite{DonahueICML14Decaf,SunAAAI16Return,CsurkaTASKCV16Unsupervised,SaxenaTASKCV16Heterogeneous}. 
Note however that the gain  is often relatively small and significantly lower than
the gain obtained with the same methods when used with the SURFBOV features.

\noindent {\bf Fine-tuning deep CNN architectures.}
The second and most used solution is to fine-tune the deep network model on the new type of 
data and for the new task \cite{ZeilerX13Visualizing,OquabCVPR14Learning,BabenkoECCV14Neural,ChuTASKCV16Best}. 
But fine-tuning requires  in general a relatively large amount of annotated data which  is not available 
 for the target domain, or it is  very limited. Therefore, the model is in general fine-tuned on the source 
- augmented with, when available,  the few labeled target instances -
which allows in a first place to adjust the deep model to the new task\footnote{This is done by 
replacing the class prediction layer to correspond to the new set of classes.}, 
common between the source and target in the case of DA. 
This is fundamental if the targeted classes do not belong to the classes used to pretrain  the deep model.  
However, if the domain difference between the source and  target is important, 
fine-tuning the model on the source might  over-fit the model for the source. In this case  
the performance of the fine-tuned model on the target data can be worse than just training the class prediction layer 
or as above,  using the model as feature extractor and training a classifier\footnote{Note that the
two approaches are  equivalent when the layer preceding the class prediction layer are extracted.}
 with the  corresponding DeCAF features \cite{ChopraWREPL13Dlid,SunAAAI16Return}.

\begin{figure}[ttt]
\begin{center}
\includegraphics[width=.95\textwidth]{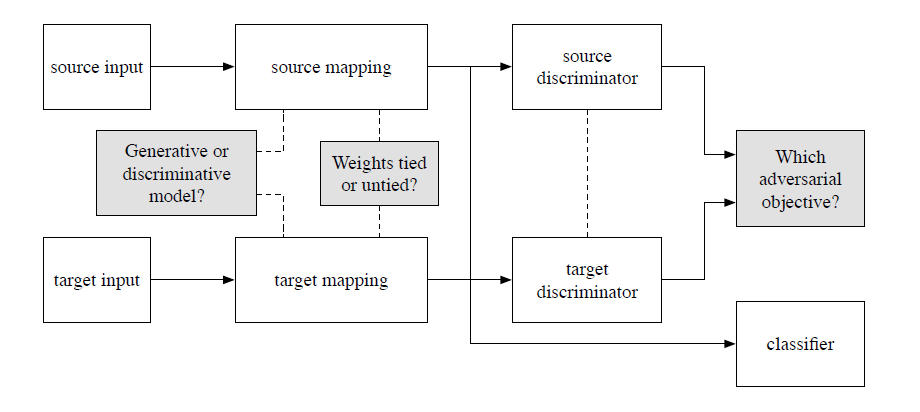}
\caption{Adversarial adaptation methods can be viewed as instantiations of the same framework
with different choices regarding their properties~\cite{TzengWAT16Adversarial}  (Image courtesy E. Tzeng).} 
\label{fig.AAD}
\end{center}
\end{figure}

\subsection{DeepDA architectures}
Finally,  the most promising are the deep domain adaptation (deepDA)
methods that are based on deep learning architectures designed for domain   adaptation.
One of the first deep model used for DA  is the Stacked Denoising 
Autoencoders \cite{VincentICML08Extracting} proposed 
to adapt  sentiment classification between reviews of different  products \cite{GlorotICML11Domain}.  
This model  aims at  finding common features
between the source and target collections relying on denoising autoencoders. This is done by 
training a multi-layer  neural network   to reconstruct input data from partial random
corruptions  with backpropagation. The Stacked Marginalized Denoising 
Autoencoders \cite{ChenICML12Marginalized} (see also in Section \ref{sec.homogeneous}) 
is a variant of the SDA,  where  the random corruption is marginalized out and hence yields a 
unique optimal solution (feature transformation) computed in
closed form  between layers. 

The  Domain Adaptive Neural Network\footnote{Code available at \url{https://github.com/ghif/mtae}}
 \cite{GhifaryICCV15Domain} uses such denoising auto-encoder 
as a pretraining stage.  To ensure that the model  pretrained on the source 
continue to adapt to the target, the  MMD  is  embedded 
as a regularization in the supervised backpropagation process (added to the cross-entropy based classification 
loss of the labels source examples).

The Deep Learning for Domain Adaptation \cite{ChopraWREPL13Dlid}, inspired by the 
intermediate representations on the  geodesic path \cite{GongCVPR12Geodesic,GopalanPAMI14Unsupervised},
proposes a deep model based interpolation between  domains. This is achieved by a deep nonlinear feature extractor 
trained in an unsupervised manner using the Predictive Sparse Decomposition \cite{KavukcuogluX10Fast}
on intermediate datasets,  where the amount of source data is  gradually replaced by  target samples. 

\cite{AljundiTASKCV16Lightweight} proposes a light-weight domain adaptation method,
which, by using only a few target samples,  analyzes and reconstructs the output of the filters that were found 
affected by the domain shift.  The aim of the reconstruction is to make the filter responses 
given a target image resemble 
to the  response map of a source image.  This is done by  simultaneously 
selecting  and  reconstructing  the response maps of the bad filters  
 using a Lasso based optimization with a KL-divergence measure that guides the filter selection process. \\

Most  DeedDA methods follow a Siamese architectures \cite{Bromley93IJPRAISignature} with two streams, 
representing the source and target models (see for example Figure \ref{fig.AAD}), 
 and  are  trained with a combination of a {\em classification loss} and a 
{\em discrepancy loss} \cite{TzengX14Deep,LongICML15Learning,GhifaryICCV15Domain,SunTASKCV16Deep,LongX16Deep}
or  an {\em adversarial loss}. The classification loss depends 
on the labeled source data.    The  discrepancy loss  aims to diminish the shift  
between the two domains  while the adversarial loss tries to encourage a common feature space through 
an adversarial objective with respect to a domain discriminator.   \\

\noindent {\bf Discrepancy-based methods.} These methods,   
inspired by  the shallow feature space transformation approaches 
described in Section \ref{sec.homogeneous}, uses 
in  general  a discrepancy based on  MMD  defined between corresponding activation layers 
of the two streams of the Siamese  architecture.  One of the first such method 
is the Deep Domain Confusion (DDC)~\cite{TzengX14Deep} where the layer to be considered 
for the discrepancy and its dimension is  automatically selected amongst a set of  fine-tuned  
networks  based on linear MMD between   the source and the target. Instead of using a
single layer and  linear MMD,    Long \etal proposed the Deep Adaptation 
Network\footnote{Code available at \url{https://github.com/thuml/transfer-caffe}} (DAN)~\cite{LongICML15Learning}  
that consider  the sum of MMDs defined between  several layers,   including the soft prediction layer too. 
Furthermore, DAN  explore multiple kernels  for adapting these deep representations,
which substantially enhances adaptation effectiveness
compared to a single kernel method used in \cite{GhifaryICCV15Domain} and \cite{TzengX14Deep}.   
This was further improved  by  the  Joint Adaptation Networks~\cite{LongX16Deep}, which 
instead of  the sum of  marginal distributions (MMD) defined between  different layers, 
consider the joint  distribution discrepancies of these features.

\begin{figure}[ttt]
\begin{center}
\includegraphics[width=.8\textwidth]{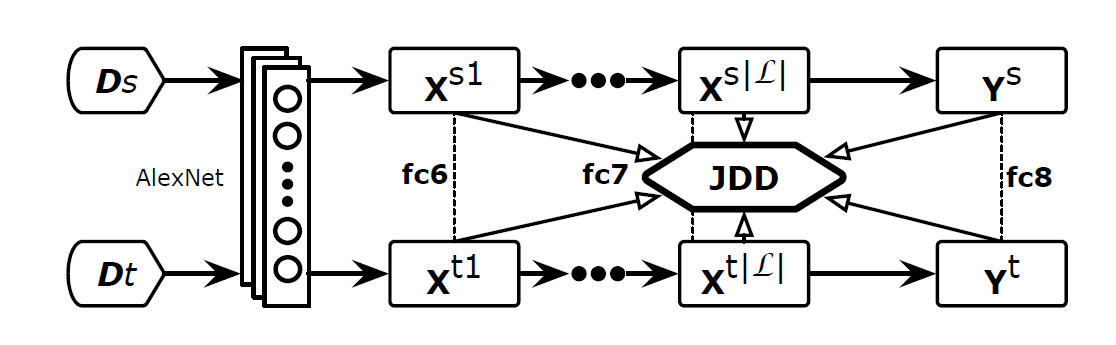}
\caption{The JAN~\cite{LongX16Deep} minimizes a joint distribution discrepancy of 
several intermediate layers including the  soft prediction one. (Image courtesy M. Long).} 
\label{fig.JDD}
\end{center}
\end{figure}

The Deep CORAL~\cite{SunTASKCV16Deep}  extends 
the shallow CORAL~\cite{SunAAAI16Return}  method  described in Section~\ref{sec.shallow}
to deep architectures\footnote{Code available at \url{https://github.com/VisionLearningGroup/CORAL}}. 
The main idea is to
learn a nonlinear transformation that aligns correlations of activation layers between the two streams. 
This idea is similarly to DDC and DAN except that instead of MMD 
the CORAL loss\footnote{Note that this loss  can be seen as
minimizing the MMD with a polynomial kernel.}
 (expressed by the distance between the covariances) 
is used to minimize discrepancy between the domains.

In contrast to the above methods,  Rozantsev \etal~\cite{RozantsevX16Beyond} consider the  MMD 
between the weights of the source respectively target models of different  layers, where an extra 
regularizer term ensures that the  weights in the two models remains linearly related. \\

\begin{figure}[ttt]
\begin{center}
\includegraphics[width=1\textwidth]{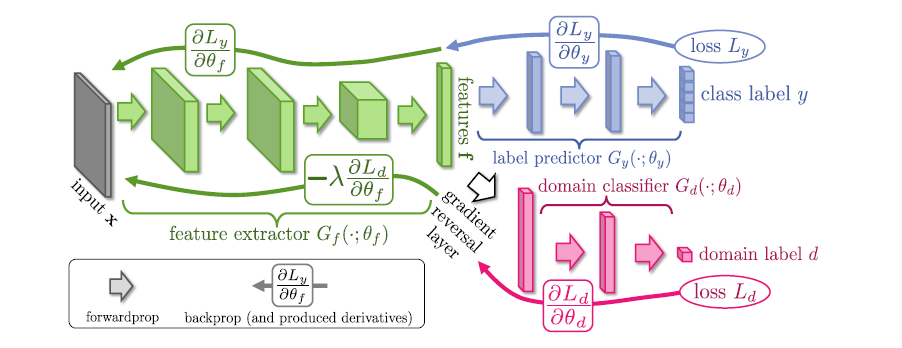}
\caption{The DANN  architecture including a feature extractor (green) and a label
predictor (blue), which together form a standard feed-forward architecture. Unsupervised
DA is achieved by the gradient reversal layer  that multiplies
the gradient by a certain negative constant during the backpropagation-based training to
ensures that the feature distributions over the two domains are made indistinguishable.
 (Image courtesy Y. Ganin~\cite{GaninJMLR16Domainadversarial}).} 
\label{fig.DANN}
\end{center}
\end{figure}

\noindent {\bf  Adversarial discriminative models.} 
The aim of these  models is  to encourage domain confusion through 
an adversarial objective with respect to a domain discriminator. \cite{TzengWAT16Adversarial}
proposes  a unified view of existing adversarial DA methods by comparing them depending 
on the loss type, the weight sharing strategy  between the two streams and,  on whether they are
discriminative or generative (see illustration in Figure \ref{fig.AAD}).
 Amongst the discriminative models we have 
the model proposed in \cite{TzengICCV15Simultaneous} using a confusion loss, 
the Adversarial Discriminative Domain Adaptation \cite{TzengWAT16Adversarial} that
considers an inverted label GAN loss \cite{GoodfellowNIPS14Generative} and 
the Domain-Adversarial Neural Network \cite{GaninJMLR16Domainadversarial} with a minimax loss. 
The generative methods, additionally to the discriminator,  relies on a generator,  which,   
 in general, is a Generative Adversarial Network (GAN) \cite{GoodfellowNIPS14Generative}.

The domain confusion based model\footnote{Code available at \url{https://github.com/erictzeng/caffe/tree/confusion}} 
proposed in \cite{TzengICCV15Simultaneous} considers  a
domain confusion objective, under which the mapping is trained with both unlabeled and sparsely labeled target data
using a cross-entropy loss function against a uniform distribution.  
The model simultaneously optimizes  the domain invariance to facilitate domain
transfer and uses a soft label distribution matching loss to transfer information between tasks.

The Domain-Adversarial Neural Networks\footnote{Code available at \url{https://github.com/ddtm/caffe/tree/grl}}
 (DANN)~\cite{GaninJMLR16Domainadversarial}, 
 integrates  a gradient reversal layer into the standard architecture to
promote the emergence of features that are  discriminative for the main learning task 
on the source domain and  indiscriminate with respect to the shift between the domains (see Figure ~\ref{fig.DANN}). 
This layer is left unchanged during  the forward propagation and its  gradient reversed during 
backpropagation.

The Adversarial Discriminative Domain Adaptation \cite{TzengWAT16Adversarial} uses 
 an inverted label GAN loss to split the optimization into two
independent objectives, one for the generator and one for the discriminator. In contrast
 to the above methods, this model  considers independent source and target mappings  
 (unshared weights between the two streams)  allowing 
domain specific feature extraction to be learned, where  the target 
weights are initialized by the network pretrained on the source.   \\

\noindent {\bf Adversarial generative models.}
These models combine the discriminative model with  a generative component
in general based on GANs \cite{GoodfellowNIPS14Generative}.  As such,  the 
Coupled Generative Adversarial Networks \cite{LiuNIPS16Coupled} 
consists of a tuple of GANs  each   corresponding to  one of the domains.
It learns a joint distribution of multi-domain images and
enforces a weight sharing constraint to limit the network capacity.

The model proposed  in \cite{BousmalisX16Unsupervised} also exploit
 GANs  with the aim to generate source-domain images such that they appear as if they were drawn
from the target domain.   Prior knowledge regarding the
low-level image adaptation process, such as foreground-background segmentation mask,
can be integrated in the model through content-similarity loss defined by a
masked Pairwise Mean Squared Error \cite{EigenNIPS14Depth} between 
the unmasked pixels of the source and generated images.
As the model decouples the process of domain adaptation
from the task-specific architecture, it is able to generalize also to object classes unseen during the
training phase. \\

\begin{figure}[ttt]
\begin{center}
\includegraphics[width=0.9\textwidth]{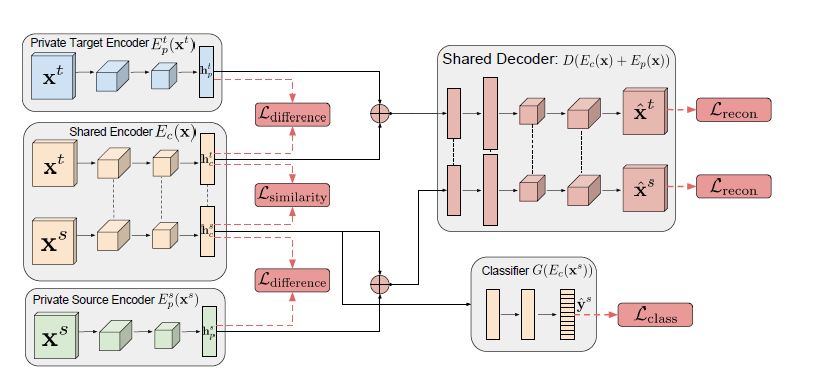}
\caption{The DSN  architecture combines shared and domain specific encoders, 
which learns common and  domain specific 
 representation components respectively with a shared decoder that learns to reconstruct the input  samples.  
 (Image courtesy K. Bousmalis~\cite{BousmalisNIPS16Domain}).} 
\label{fig.DSN}
\end{center}
\end{figure}

\noindent {\bf Data reconstruction (encoder-decoder) based methods.}
In contrast to the above methods, the Deep Reconstruction Classification 
Network\footnote{Code available at \url{https://github.com/ghif/drcn}}  proposed in ~\cite{GhifaryECCV16Deep} 
combines the standard convolutional network for source label prediction 
with a deconvolutional network~\cite{ZeilerCVPR10econvolutional} for target data reconstruction.
To  jointly learn  source label predictions and  
unsupervised target data reconstruction, the model  alternates between 
unsupervised and supervised  training.
The  parameters of the encoding are shared across both tasks, 
while the decoding parameters are separated.  The data reconstruction can be
viewed as an auxiliary task to support the adaptation of the label prediction.  

The Domain Separation Networks (DSN) \cite{BousmalisNIPS16Domain} 
introduces the notion of a private  subspace for each domain, which captures domain specific
properties, such as background and low level image statistics. A shared subspace, enforced through
the use of autoencoders and explicit loss functions, captures common features between the domains. 
The model integrates  a reconstruction loss using a shared decoder, which learns to reconstruct 
the input sample by using both the private (domain specific) 
and source representations (see  Figure~\ref{fig.DSN}). \\

\noindent {\bf Heterogeneous deepDA.}
Concerning heterogeneous or multi-modal deep domain adaptation, we can mention the 
 Transfer Neural Trees \cite{ChenECCV16Transfer} proposed  to relate heterogeneous cross-domain data. 
It is a two stream network, one stream for each modality, where 
 the weights in the latter stages of the network are shared.  As the prediction layer, 
 a Transfer Neural Decision Forest (Transfer-NDF) is used that performs jointly
adaptation and classification. 

\begin{figure}[ttt]
\begin{center}
\includegraphics[width=0.9\textwidth]{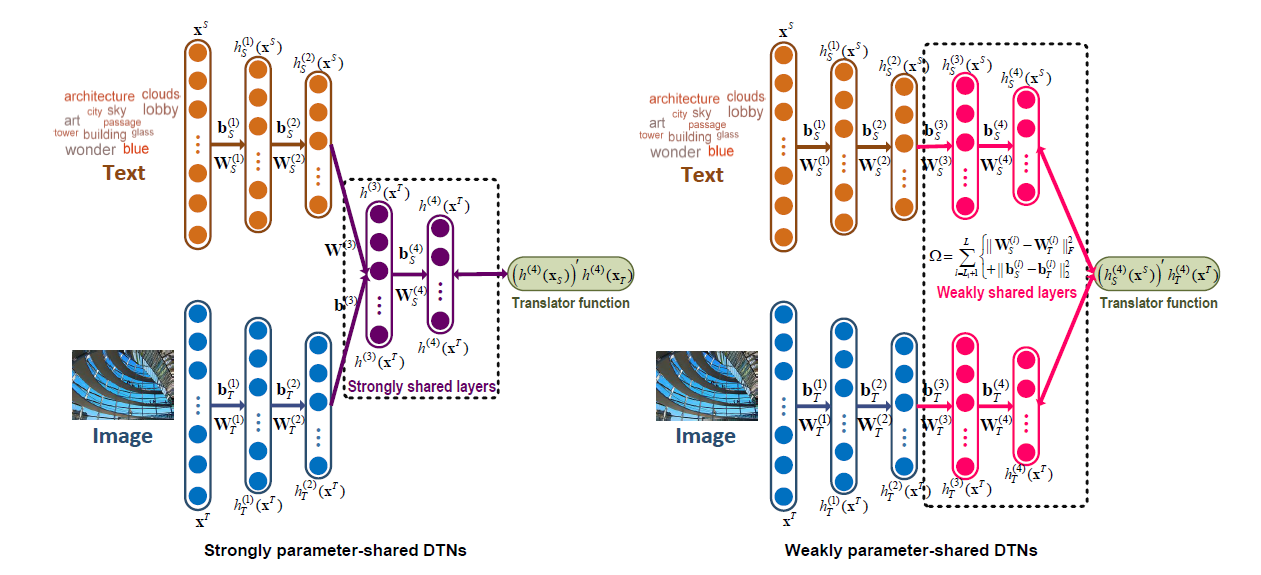}
\caption{The DTN  architecture with strongly-shared and weakly-shared  parameter layers. 
 (Image courtesy X. Shu~\cite{ShuACMMM15Weaklyshared}).} 
\label{fig.DTN}
\end{center}
\end{figure}

The weakly-shared Deep Transfer Networks for
Heterogeneous-Domain Knowledge Propagation \cite{ShuACMMM15Weaklyshared}  learns a 
domain translator function from  multi-modal source data that can be used to 
predict class labels in the target even if only one of the modality is present.
The proposed structure has the advantage to be flexible enough to represent
both domain-specific features and  shared features across domains
(see  Figure~\ref{fig.DTN}).

\section{Beyond image classification}
\label{sec:beyondclassif}

In the previous sections, we attempted to provide an overview of visual  DA methods 
with  emphasis on image categorization. Compared to this vast literature
focused on object recognition, 
relatively few papers  go beyond image classification and address 
domain adaptation  related to other 
computer vision problems such as  object  detection, semantic segmentation,
pose estimation,  video event or  action detection. One of the main reason  is 
probably due to the fact that these problems are more complex and have often 
additional challenges and requirements (\eg precision related to the localization 
in the case of detection, pixel level accuracy required for image segmentation, 
increased amount of annotation burden needed for videos, \etc.)  Moreover, 
 adapting visual representations such as  contours,
deformable and articulated 2-D or 3-D models, graphs, random fields  or 
visual dynamics, is less obvious  with classical  {\em vectorial  DA} techniques.

Therefore,  when  these tasks are addressed in the context of domain adaptation, 
 the problem is generally rewritten  as a classification problem with 
 vectorial feature representations  and a set of predefined class labels.  In this case 
 the main challenge  becomes  finding the best vectorial representation for the given the task. 
 When this is possible, shallow DA methods, described in the Section \ref{sec.shallow},
 can be applied to the problem.
Thereupon,  we can find in the literature  DA solutions such as Adaptive SVM \cite{YangMM07Crossdomain},  
DT-SVM \cite{DuanCVPR09Domain},
A-MKL \cite{DuanCVPR12Exploiting} or Selective Transfer Machine \cite{ChuCVPR13Selective}
 applied to video concept detection \cite{YangICCV13Crossdomain}, video  
event recognition \cite{DuanCVPR12Exploiting},
 activity recognition \cite{FarajidavarBMVC12Domain,ZhuBMVC13Enhancing}, 
 facial action unit detection \cite{ChuCVPR13Selective},
and 3D Pose Estimation \cite{YamadaECCV12OBias}.

When rewriting the problem into classification of  vectorial representation 
is less obvious, as in the case of image segmentation, 
where the output is  a  structured output, or detection where 
the output is a set of  bounding boxes,  most often the target training set is simply 
augmented with the source data and traditional - segmentation, detection, \etc. - methods are used. 
To overcome the lack of labels in the  target domain,  source data is often gathered by  crawling the Web 
(webly supervised) \cite{DivvalaCVPR14Learning,ChenICCV15Webly,CrowleyCVAA16Art} or 
the target set is enriched with synthetically generated  data. 
The usage of the synthetic data  became even  more  popular since  
the massive adoption of deep CNNs to perform computer vision tasks requiring large amount of annotated data.     \\

\begin{figure}[ttt]
\begin{center}
\includegraphics[width=0.8\textwidth]{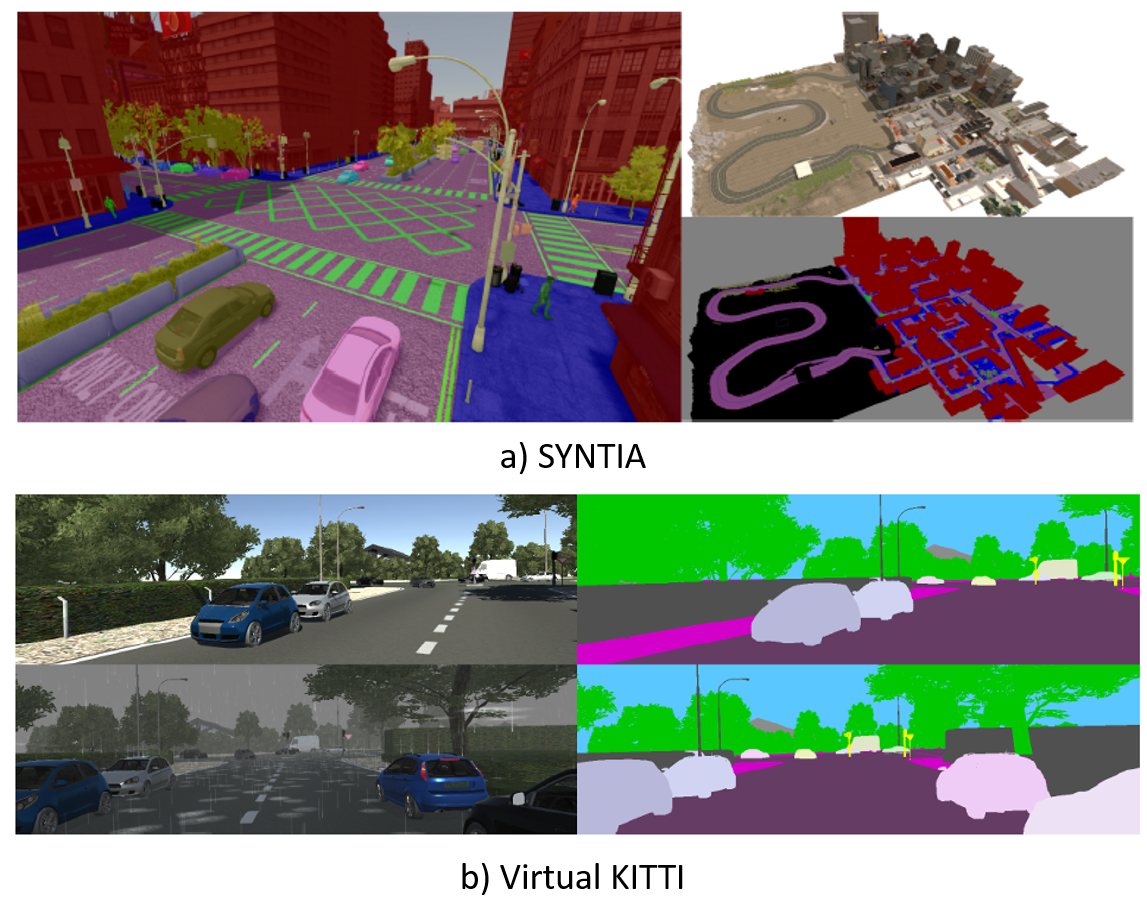}
\caption{Virtual word examples: SYNTHIA (top), Virtual KITTI (bottom).} 
\label{fig.virtualWorld}
\end{center}
\end{figure}

\noindent {\bf Synthetic data based adaptation.}
Early methods use 3D CAD models to improve solutions for pose and viewpoint 
estimation \cite{AgarwalACCV06Local,ShottonCVPR11Realtime, PanaredaBustoBMVC15Adaptation, SuICCV15Render},
object and object part  detection \cite{StarkBMVC10Back, PepikCVPR12Teaching, SunBMVC14Virtual,RozantsevCVIU15Rendering,PengICCV15Learning,HattoriCVPR15Learning,MassaCVPR16Deep,BochinskiAVSS16Training},
segmentation  and  scene understanding \cite{SatkinBMVC12Data,ChenCVPR14Beat, PaponICCV15Semantic}. 
The recent progresses in computer graphics and modern high-level
generic graphics platforms such as game engines enable to generate photo-realistic 
virtual worlds with diverse, realistic, and physically plausible events and actions. 
Popular virtual words are 
SYNTHIA\footnote{Available at \url{http://synthia-dataset.net}} \cite{RosCVPR16Synthia}, 
Virtual KITTI\footnote{Available at\url{http://www.xrce.xerox.com/Research-Development/Computer-Vision/Proxy-Virtual-Worlds}}~\cite{GaidonCVPR16Virtual}
and GTA-V \cite{RichterECCV16Playing}  (see also Figure \ref{fig.virtualWorld}). 

Such virtually generated and controlled environments come with different levels of labeling for free
and therefore have great promise for deep learning across a variety
of computer vision problems, including optical flow \cite{MeisterCEMT11Real,ButlerECCV12Naturalistic,OnkarappaMTA15Synthetic,MayerCVPR16Large},
object trackers \cite{TaylorCVPR07OVVV,GaidonCVPR16Virtual}, depth estimation from RGB \cite{ShafaeiBMVC16Play},
object detection \cite{MarinCVPR10Learning, VazquezPAMI14Virtual,XuITS14Learning}
semantic segmentation \cite{HandaCVPR16Understanding, RosCVPR16Synthia,RichterECCV16Playing}
or human actions recognition \cite{SouzaX16Procedural}.

\begin{figure}[ttt]
\begin{center}
\includegraphics[width=0.95\textwidth]{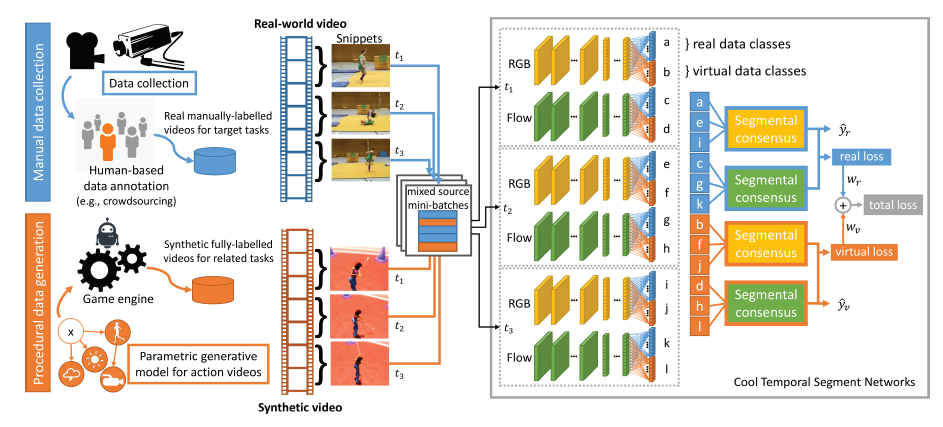}
\caption{Illustration of the Cool-TSN deep multi-task learning architecture~\cite{SouzaX16Procedural}
 for end-to-end action recognition in videos. (Image courtesy C. De Souza).}
\label{fig.CoolTSN}
\end{center}
\end{figure}
 
In most cases, the synthetic data is used to enrich the real data for building 
the models.  However, DA techniques can further help to 
adjust the model trained with virtual data (source) to real data (target)
especially when no or few labeled examples are available 
in the real domain \cite{TzengX15Towards,XuIJCV16Hierarchical,RosCVPR16Synthia,SouzaX16Procedural}.
As such, \cite{TzengX15Towards}  propose a deep spatial feature point architecture
for  visuomotor representation  which, using synthetic examples and  a few supervised examples,
transfer the pretrained model to  real imagery. This is done by combining  
a pose estimation loss, a domain confusion loss that aligns 
the synthetic and real domains, and a contrastive loss that aligns specific pairs in the feature space. 
 All together, these three losses ensure that the representation is suitable to the pose estimation 
task while remaining robust to the synthetic-real domain shift.

The Cool Temporal Segment Network~\cite{SouzaX16Procedural} is an end-to-end action recognition model
for real-world target categories that combines a few examples of labeled real-world videos with a large number of
procedurally generated synthetic videos.  The model uses a deep multi-task representation learning 
architecture,  able to mix synthetic
and real videos even if the action categories differ between the real and synthetic sets (see Figure~\ref{fig.CoolTSN}).

\begin{figure}[ttt]
\begin{center}
\includegraphics[width=0.9\textwidth]{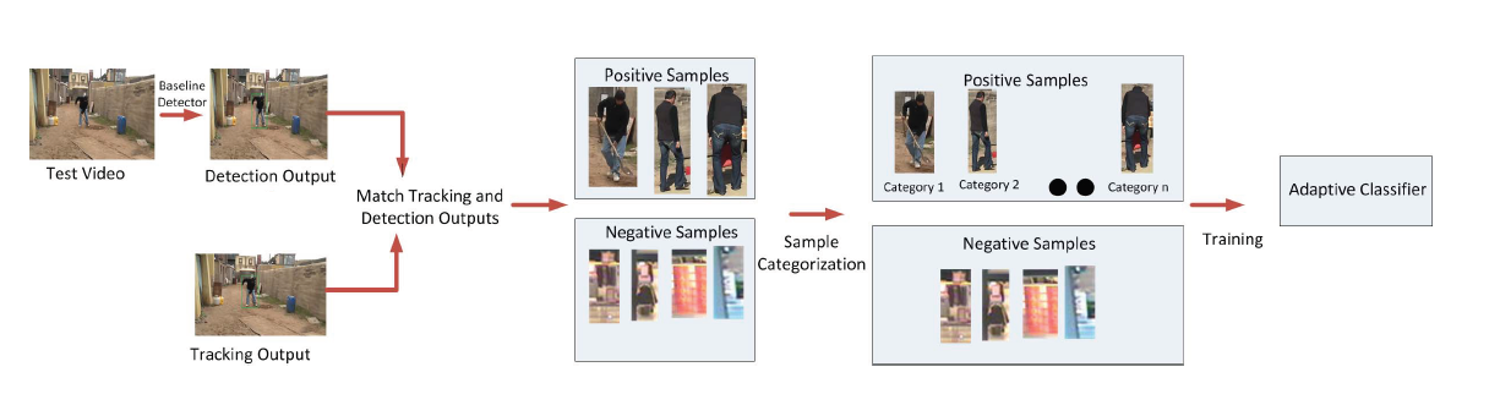}
\caption{Online adaptation of the generic detector with tracked regions. (Image courtesy P. Sharma~\cite{SharmaCVPR13Efficient}).} 
\label{fig.onlineadapt}
\end{center}
\end{figure}

 \subsection{Object detection}
 \label{sec:objdet}

Concerning visual applications, after the  image level categorization task,  object detection 
received the most attention from the visual DA/TL community. Object detection models,  until recently,  
 were composed of  a window selection mechanism and appearance based classifiers 
trained on the features extracted from  labeled bounding boxes.
At test time,  the classifier  was used to decide if a region of interest 
obtained by  sliding windows or generic   
window selection models \cite{WangICCV13Regionlets,ZitnickECCV14Edge,UijlingsIJCV13Selective}
 contains  the object or not.

 Therefore,  considering the  window selection mechanism  as 
 being domain independent,  standard DA methods can be integrated with  the  appearance based classifiers
 to adapt to the target domain the models trained on the source domain.  The Projective Model Transfer SVM (PMT-SVM)
and the Deformable Adaptive SVM  (DA-SVM) proposed in \cite{AytarICCV11Tabula} are such methods,  which
adapt  HOG deformable source templates \cite{DalalCVPR05Histograms,FelzenszwalbPAMI10Object}
with  labeled target bounding boxes (SS scenario),  and the adapted  template is used  at test time to detect the  
presence or absence of an object class in sliding windows. In \cite{DonahueCVPR13SemiSupervised}
the PMT-SVM was  further  combined  with  MMDT \cite{HoffmanICLR13Efficient}  to handle complex domain shifts.  
The detector is  further improved by a smoothness constraints imposed on the classifier scores 
 utilizing instance correspondences (\eg  the same object observed simultaneously from multiple
views or tracked between video frames). 

\cite{MirrashedWACV13Domain} uses the
 TCA \cite{PanTNN11Domain} to adapt  image level HOG representation 
 between source and target  domains for object detection.  \cite{ZhangCVPR08Taylor} proposes
a Taylor Expansion Based Classifier Adaptation for either boosting or logistic regression to 
adapt person detection between videos acquired in  different meeting rooms. \\

\noindent {\bf Online adaptation of the detector.}
Most early works related to object detector adaptation  concern  online adaptation of a generic  detector
trained on strongly labeled images (bounding boxes) to detect objects (in general cars or pedestrians) in 
videos. These methods exploit redundancies in videos  to obtain prospective positive target examples (windows) 
either by background  modeling/subtraction \cite{RothCVPR09Classifier, StalderPETS09Exploring}, or by 
combination of object tracking with regions proposed by the generic detector \cite{TangNIPS12Shifting,SharmaCVPR13Efficient,GaidonX14Selflearning,GaidonBMVC15Online}
(see the main idea in Figure \ref{fig.onlineadapt}).
Using these designated target samples in the new frame the model is updated involving  semi-supervised approaches
such as self-training \cite{RosenbergWACV05Semisupervised,WuCVPR07Improving} 
or co-training \cite{JavedCVPR05Online,LevinICCV13Unsupervised}. 

\begin{figure}[ttt]
\begin{center}
\includegraphics[width=0.85\textwidth]{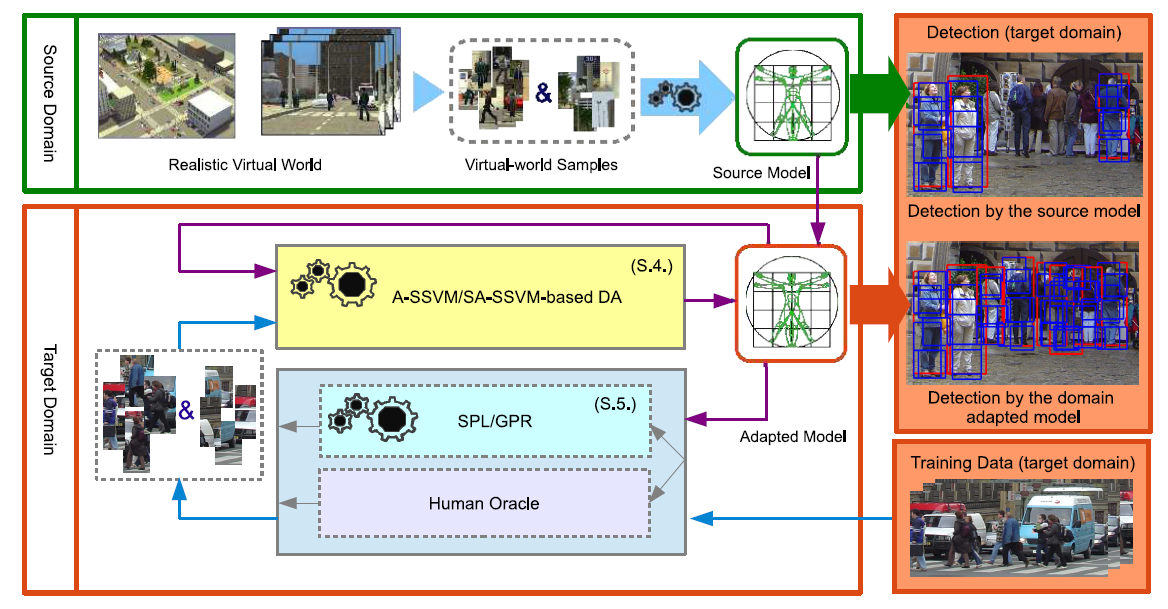}
\caption{\small Domain Adaptation of DPM based on SA-SSVM~\cite{XuPAMI14Domain} (Image courtesy J. Xu).}
\label{fig:SA-SSVM}
\end{center}
\end{figure}

For instance, \cite{WangCVPR12Detection} proposes
a non-parametric detector adaptation algorithm, which adjusts an offline frame-based object detector to the
visual characteristic of a new  video clip.  The  Structure-Aware Adaptive Structural SVM (SA-SSVM) \cite{XuPAMI14Domain}
 adapts online the deformable part-based model \cite{DollarCVPR09Pedestrian} for pedestrian detection
 (see Figure \ref{fig:SA-SSVM}). To handle the case when no target label is available, a strategy
inspired by self-paced learning  and supported by a Gaussian Process Regression is used 
to  automatically label samples in the target domains.
The temporal structure of the video is exploited through similarity constraints imposed on the adapted detector.  \\

\noindent {\bf Multi-object tracking.}
Multi-object tracking  aims at automatically detecting and tracking individual object (\eg car or pedestrian)
instances \cite{SharmaCVPR12Unsupervised,GaidonX14Selflearning,GaidonBMVC15Online}.
These methods  generally capitalizes on multi-task and multi-instance learning to perform 
category-to-instance adaptation. For instance,   \cite{SharmaCVPR12Unsupervised} introduces
a Multiple Instance Learning (MIL) loss function for Real Adaboost,  which is used within 
a tracking based  unsupervised online sample collection mechanism to  incrementally adjust the  
pretrained detector.

\cite{GaidonX14Selflearning} propose an unsupervised, online and self-tuning learning algorithm to optimize
a multi-task learning based  convex objective
involving a high-precision/low-recall off-the-shelf generic detector. The method exploits
the data structure to jointly learn an ensemble of instance-level
trackers, from which  adapted category-level object detectors  are derived. 
The main idea in \cite{GaidonBMVC15Online} is to  jointly learn all detectors 
(the target instance models and the generic one)
using an online adaptation via Bayesian filtering coupled with  multi-task learning to efficiently share
parameters and reduce drift, while gradually improving recall.

The transductive approach  in \cite{TangNIPS12Shifting} re-trains the detector with automatically
discovered target domain examples starting with the easiest first,  and  
iteratively re-weighting labeled source samples by scoring trajectory tracks.
\cite{SharmaCVPR13Efficient} introduces a  multi-class random fern adaptive classifier where 
different categories of the positive samples (corresponding to different video tracks) are considered  as different
target classes, and all negative online samples are considered as a single negative target class. 
\cite{BreitensteinPAMI11Online} proposes a particle filtering framework for
 multi-person tracking-by-detection  to predict the target locations.  \\

\noindent {\bf Deep neural architectures.} 
More recently,  end-to-end deep learning object detection models were proposed that  integrate and learn simultaneously
the  region proposals and the  object appearance.   In general, these models  are initialized 
by  deep models pretrained  with image level annotations 
(often on the ILSVRC datasets \cite{RussakovskyIJCV15Imagenet}). In fact, the pretrained deep model
combined with class-agnostic region of interest  proposal, can already be used to predict the 
presence or absence of the target object in the proposed local regions \cite{SermanetX13Overfeat,GirshickCVPR14Rich,OquabCVPR14Learning,HoffmanNIPS14Lsda}.
When strongly labeled target data is available,  the model can further be fine-tuned using the labeled 
bounding boxes to improve both the recognition and the object localization.
Thus, the Large Scale Detection through 
Adaptation\footnote{Code available at \url{https://github.com/jhoffman/lsda/zipball/master}} \cite{HoffmanNIPS14Lsda}  
learns to transform an image classifier into an object detector by  fine-tuning the CNN model, 
pretrained on images, with a set of labeled bounding boxes. 
 The advantage of this model is that it generalizes well  even for 
localization of classes  for which there were no  bounding box annotations during the training phase.

Instead fine-tuning,  \cite{RajBMVC15Subspace}  uses 
Subspace Alignment \cite{FernandoICCV13Unsupervised}  to adjust class specific  
representations  of  bounding boxes (BB) between the source and target domain. The source 
BBs are extracted from  the strongly annotated training set, while 
the target BBs are  obtained with the RCNN-detector \cite{GirshickCVPR14Rich} trained on the source set.
The detector is then re-trained with the target aligned source
features and used to classify the target data  projected into the target subspace.

\section{Beyond domain adaptation: unifying perspectives}
\label{sec:discussion}

The aim of this section  is to relate domain
adaptation to other machine learning solutions.
First in Section \ref{sec:TL} we  discuss how DA is related to 
other transfer learning (TL) techniques. 
Then,  in Section \ref{sec:ML}  we connect DA to several classical  machine learning approaches
illustrating how these methods are exploited in various DA solutions. Finally,  in Section 
\ref{sec:MM} we examine the relationship between  heterogeneous DA and  multi-view/multi-modal learning.

\subsection{DA within transfer learning}
\label{sec:TL}

As shown in Section \ref{sec.definitions},
DA is a particular case of the transductive transfer learning 
 aimed  to solve  a classification task  common to the source and target,
 by simultaneously exploiting labeled source  and unlabeled target examples
(see also Figure \ref{fig:TL}).  
As such,  DA is opposite to unsupervised TL, where both domains and tasks are
different with labels available neither for source nor for target. 

DA is also different  from self-taught  learning \cite{RainaICML07Selftaught}, 
which exploits a limited labeled target data for a classification task together  with 
a large amount of unlabeled source data  mildly related to the task.   The main idea behind self-taught learning is 
to explore  the  unlabeled source data and to discover repetitive patterns  
that could be used for the supervised learning task.  

On the other hand,  DA is more closely related  to 
domain generalization \cite{MuandetICML13Domain,XuECCV14Exploiting,GhifaryICCV15Domain,GanCVPR16Learning,NovotnyBMVC16Ihave},
multi-task learning \cite{Caruana1997MLMultitask,EvgeniouSIGKDD04Regularized,RomeraParedesICML13Multilinear}  or
few-shot learning \cite{MillerCVPR10Learning,FeiFeiPAMI06Oneshot} discussed below.\\

\noindent{\bf Domain generalization.}
Similarly to  multi-source DA \cite{ChattopadhyaySIGKDD11Multisource,DuanTNNLS12Domain,JhuoCVPR12Robust}, 
domain generalization methods \cite{MuandetICML13Domain,XuECCV14Exploiting,GhifaryICCV15Domain,GanCVPR16Learning,NovotnyBMVC16Ihave}
aim to average knowledge from several related source domains,   
in order to learn a model for a new target domain. But, in contrast to DA where 
unlabeled target instances are available to adapt the model, 
in domain generalization, no target example is  accessible 
at training time. \\

\noindent{\bf Multi-task learning.}
In multi-task learning \cite{Caruana1997MLMultitask,EvgeniouSIGKDD04Regularized,RomeraParedesICML13Multilinear} 
different tasks (\eg sets of the labels) are learned  at the same time using a shared representation such that 
 what is learned for each task can help in learning the other tasks. 
If we considering  the tasks  in DA  as  domain  source and target) specific tasks, 
a semi-supervised DA method can be seen as a sort of two-task learning problem  where, in particular, learning 
 the source specific task  helps learning the target specific task. Furthermore, in the case of 
multi-source domain adaptation \cite{MansourNIPS09Domain,DuanICML09Domain,TommasiCVPR10Safety,YaoCVPR10Boosting,SunNIPS11Twostage,HoffmanECCV12Discovering,ShekharCVPR13Generalized,GongNIPS13Reshaping,GopalanPAMI14Unsupervised,CsurkaTASKCV14Domain}
 different source specific tasks  are jointly exploited in the interest of the target task.

On the other hand,  as we have seen in Section \ref{sec:objdet}, 
multi-task learning techniques can be  beneficial  for online DA, in particular  
 for  multi-object tracking and detection \cite{GaidonX14Selflearning,GaidonBMVC15Online}, 
 where the generic object detector (trained on source data) 
 is adapted for each individual object instance. \\

\noindent{\bf Few-shot learning.}
Few-shot learning \cite{MillerCVPR10Learning,FeiFeiPAMI06Oneshot,TommasiCVPR10Safety,TommasiBMVC09EMore}   
aims to learn information about  object categories when  only a few training images  are available for training. 
This is done by   making use of prior knowledge of related  categories for which larger amount of annotated 
data is available.  Existing solutions are the 
 knowledge transfer through the reuse of model parameters \cite{FinkNIPS04Object},  methods  sharing parts or 
 features \cite{BartCVPR05Crossgeneralization} or approaches relying on  contextual information \cite{MurphyNIPS03Using}. 
 
 An extreme case of few-shot learning is the {\em zero-shot learning} \cite{FerrariNIPS07Learning,LampertCVPR09Learning},  where 
the  new task is deduced  from previous tasks  without using any training data for the current task.  
To address zero-shot learning, the methods rely  either on 
nameable image characteristics  and semantic concepts \cite{FerrariNIPS07Learning,LampertCVPR09Learning,PalatucciNIPS09Zero,FuECCV14Transductive}, 
or on latent topics discovered by the system  directly from the data \cite{SharmanskaECCV12Augmented,LayneBMVC14Reid,FuPAMI14Learning}.
In both cases, detecting these attributes can be seen as the common tasks between the training classes 
(source domains) and the new classes (target domains).  \\

\noindent{\bf Unified DA and TL models.}
We have seen that the particularity of DA is the shared label space, in contrast to more generic 
TL  approaches where the focus is  on the task transfer between classes. 
However, in \cite{PatriciaCVPR14Learning} it is claimed that 
task transfer  and domain shift can be seen as different declinations of {\it learning to
learn} paradigm, \ie the ability to leverage  prior knowledge when
attempting to solve a new task. Based on this observation, 
 a common  framework  is proposed  to leverage source data regardless of the origin of the distribution mismatch.  
 Considering prior models as experts, the original features are
augmented with  the output confidence values of the source models and  target classifiers are then 
learned with these features.

Similarly, the Transductive Prediction Adaptation (TPA) \cite{ClinchantACL16Transductive} 
augments the  target features with class predictions from source experts, 
before applying the MDA framework \cite{ChenICML12Marginalized} 
on these augmented features. It is shown that MDA,  exploiting the correlations 
between the target features and source predictions,  can denoise the class predictions  and
improve classification accuracy.
In contrast to the method in \cite{PatriciaCVPR14Learning}, TPA works also in the case when no label 
is available in the target domain (US scenario).

The Cross-Domain Transformation \cite{SaenkoECCV10Adapting}  learns
a regularized non-linear transformation using  supervised data from both domains 
to map  source examples closer to the  target ones.  It is shown that the models  built
in this new  space  generalize well not only to new samples from categories used to 
train the transformation  (DA) but also 
to new categories that were not present at training time (task transfer).  
The Unifying Multi-Domain Multi-Task Learning \cite{YangICLR15Unified},  
is a Neural Network  framework that can be flexibly applied to multi-task, multi-domain 
and zero-shot learning and even to zero-shot domain adaptation.

\subsection{DA related to traditional ML methods}
\label{sec:ML}

\noindent{\bf Semi-supervised learning.}
DA can  be seen as a particular case of the semi-supervised learning \cite{ChapelleB06Semisupervised,ZhuB09Introduction},  
where, similarly to the majority of DA  approaches, unlabeled data is exploited to remedy the lack of labeled data. 
 Hence, ignoring the domain shift, traditional semi-supervised learning can be used  as a solution for  DA, 
 where the source instances form  the supervised part, and the
 target domain provides the unlabeled data. For this reason, 
 DA  methods often exploit or  
 extend  semi-supervised  learning techniques
such as transductive SVM \cite{BruzzonePAMI10Domain}, 
self-training \cite{RosenbergWACV05Semisupervised,WuCVPR07Improving,TommasiICCV13Frustratingly,CsurkaTASKCV14Domain}, 
or co-training \cite{JavedCVPR05Online,LevinICCV13Unsupervised}. When the domain shift is small, 
traditional semi-supervised methods can already bring a significant improvement over  baseline
methods obtained with the pretrained source model \cite{BruzzonePAMI10Domain}.  \\

\noindent{\bf Active learning.}
Instance selection  based DA methods exploit  ideas from  active learning \cite{Settles10Active} 
to select  instances with best potentials to  help the training process. Thus, 
the Migratory-Logit algorithm \cite{LiaoICML05Logistic} explore,  both the target and source data
 to actively select unlabeled target samples to be added to the training sets.
\cite{ShiPKDD08Actively} describes an active learning method  for relevant target data selection and labeling, which  
 combines TrAdaBoost \cite{DaiICML07Boosting} with  standard SVM. 
\cite{NovotnyBMVC16Ihave},  (see also Chapter 15), uses active learning and DA techniques to 
generalize semantic object parts (\eg animal eyes or legs) to unseen classes  (animals). 
The methods  described in \cite{Chan07Domain,RaiALNLP10Domain,SahaECMLKDD11Active,TommasiICCV13Frustratingly,CsurkaTASKCV14Domain,WangICML14Active}
combine transfer learning and  domain adaptation with the target sample selection and automatic sample 
labeling,  based on the classifier confidence.  These new samples are then used 
to iteratively update the target models. \\

\noindent{\bf Online learning.}
 Online or sequential learning \cite{ShalevShwartzPHD07Online,BottouB98Online,ShalevShwartzFTML11Domain} 
 is strongly related to  active learning; 
in  both  cases the model is iteratively and  continuously  updated using  new data. 
However, while in active learning 
 the  data to be used for the update is actively selected, in online learning 
 generally the new data is acquired sequentially.  
Domain adaptation can be combined with online learning too. 
As an example, we  presented in Section \ref{sec:objdet} the online adaptation
 for incoming video frames of a generic object detector trained offline  on labeled image sets 
\cite{BreitensteinPAMI11Online,XuPAMI14Domain}. \cite{YanTASKCV16Online} proposes  
 online adaptation of image classifier to user generated content in  social computing applications.
 
Furthermore, as  discussed in Section \ref{sec.deep},  fine-tuning 
a    deep model \cite{ZeilerX13Visualizing,ChopraWREPL13Dlid,OquabCVPR14Learning,BabenkoECCV14Neural,ChuTASKCV16Best,SunAAAI16Return},
pretrained on ImageNet (source),  for a new dataset (target),  
 can be seen as sort of semi-supervised domain adaptation. 
Both, fine-tuning as well as training 
deepDA models \cite{GaninJMLR16Domainadversarial,LongICML15Learning,GhifaryECCV16Deep}, use
sequential learning where data  batches  are used to perform the  stochastic gradient updates. 
If we assume that these batches contain the target data acquired
sequentially,  the  model learning process can be  directly used for online DA adaptation of the original model. \\

\noindent{\bf Metric learning.}
In  Section \ref{sec.shallow} we presented several  metric learning based DA methods \cite{ZhaIJCAI09Robust,SaenkoECCV10Adapting,ZhangSIGKDD10Transfer,KulisCVPR11What,CsurkaTASKCV14Domain}. 
where class labels from both domains are exploited 
to bridge the relatedness between the source and target.  Thus, \cite{ZhaIJCAI09Robust} proposes a 
new distance metric for the target domain by using the existing distance metrics 
learned on the source domain. \cite{SaenkoECCV10Adapting} uses information-theoretic 
metric learning \cite{DavisICML07Informationtheoretic} as  a distance metric across different 
domains, which was extended to 
non-linear kernels in  \cite{KulisCVPR11What}.   
\cite{CsurkaTASKCV14Domain} proposes a  metric learning adapted to the DSCM classifier, 
 while \cite{ZhangSIGKDD10Transfer} defines a multi-task metric learning framework
 to learn  relationships between  source  and target tasks. 
\cite{SaxenaTASKCV16Heterogeneous} explores various metric learning 
approaches to align deep features extracted from RGB and NIR face images.  \\

\noindent{\bf Classifier ensembles.}
Well studied in ML, classifier ensembles have also been considered for DA and  TL. As such, 
\cite{KamishimaICDM09Trbagg} applies a bagging approach for transferring the learning capabilities
of a model to different domains where a high number of trees is learned on 
both source and target data in order to build a pruned version of the final ensemble to avoid a negative
transfer.  \cite{RodnerDAGM09Learning} uses random decision forests 
to transfer  relevant features  between domains. The optimization
framework  in \cite{AcharyaWUTL12Transfer} takes as input several classifiers learned on the source domain as
well as the results of a cluster ensemble operating solely on the target domain,  yielding a
consensus labeling of the data in the target domain. 
Boosting was extended to DA  and TL in \cite{DaiICML07Boosting,AlstouhiPKDD11Adaptive,ZhangCVPR08Taylor,YaoCVPR10Boosting,ChidlovskiiCLEFWN14Assembling}.

\begin{figure}[ttt]
\begin{center}
\includegraphics[width=1\textwidth]{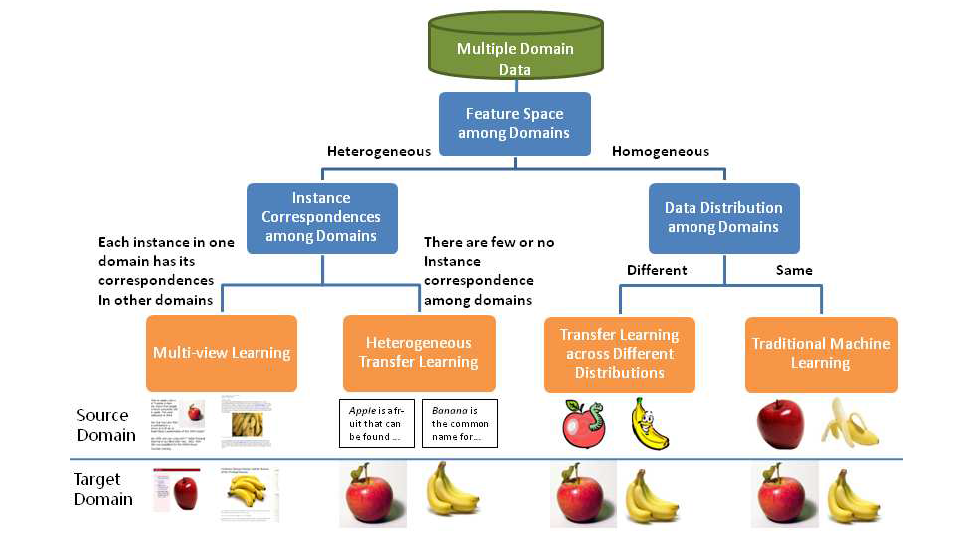}
\caption{Illustrating through an example  the difference between 
 TL to ML in the case of  homogeneous data and between 
multi-view and HTL/HDA when working with  
heterogeneous data.  Image courtesy Q. Yang~\cite{YangACL07Instance}.} 
\label{fig:heterogeneous}
\end{center}
\end{figure}

\subsection{HDA  related to multi-view/multi-modal learning}
\label{sec:MM}

In many data intensive applications,  such as video surveillance, 
social computing, medical health records or environmental sciences, 
data collected from diverse domains or obtained from various feature
extractors  exhibit heterogeneity.   For example, a person can be identified by different facets \eg
face, fingerprint,
signature or iris,  or in video surveillance, an action or event can be 
recognized  using multiple cameras.  
When working with such heterogeneous or multi-view data most, methods try to exploit simultaneously 
different modalities to build better final models. 

As such,  multi-view learning methods are related to HDA/HTL as discussed also in Section \ref{sec.heterogeneous}.
Nevertheless,  while  multi-view learning \cite{XuX13Survey,WangICML15Deep} 
assumes that  multi-view examples are available during training, 
in the case of HDA \cite{ShiICDM10Transfer,WangIJCAI11Heterogeneous,JhuoCVPR12Robust,ShekharCVPR13Generalized,YaoCVPR15Semisupervised}, 
this assumption rarely holds (see illustration  in \ref{fig:heterogeneous}).
On contrary, the aim  of HDA is to transfer information from the  source domain represented
with  one type of data (\eg text)
to the target  domain represented with  another type of data (\eg images).
While this assumption essentially differentiates the multi-view learning from HDA, 
we have  seen in Section \ref{sec.heterogeneous} that  
HDA methods often rely on  an auxiliary intermediate multi-view 
domain \cite{ZhuAAAI11Heterogeneous,QiWWW11Towards,TanSICDM14MixedTransfer,TanSIGKDD15Transitive,YangTNNLS15Learning,YanTASKCV16Online}.
Hence, HDA/HTL  can strongly benefit from  multi-view learning techniques 
such as canonical correlation analysis \cite{HardoonNC05Canonical}, 
co-training \cite{ChenNIPS11Cotraining}, spectral embedding \cite{ShiICDM10Transfer} and 
multiple kernel learning \cite{DuanPAMI12Learning}. 

Similarly to HDA/HTL relying on intermediate domains, cross-modal image retrieval methods depend
on multi-view  auxiliary  data to define 
cross-modal similarities \cite{AhPineMTA09Synthetic,JiaICCV11Learning}, or to perform 
semantic \cite{WestonML10Large,RasiwasiaACMMM10New,FromeNIPS13Devise,YanCVPR15Deep}
or multi-view embedding \cite{HardoonNC05Canonical,NgiamICML11Multimodal,SharmaCVPR12Generalized,GongIJCV14multiview,CaoX16Generalized,WangCVPR16Learning}. 
Hence, HDA/HTL  can strongly benefit from  such cross-modal data representations.

In the same spirit, {\it webly supervised}  approaches \cite{FergusICCV05Learning,WangPAMI08Annotating,SchroffICCV07Harvesting,BergamoNIPS10Exploiting,DivvalaCVPR14Learning,ChenICCV15Webly,GanECCV16WeblySupervised} 
are also related to DA and HDA as is these approaches rely 
on  collected Web  data  (source) data  used  to refine   the target model. 
As such, \cite{DuanPAMI12Visual} uses multiple kernel learning  to adapt 
visual events learned from the Web data for video clips. \cite{SunACMMM15Temporal} 
and \cite{GanECCV16WeblySupervised} propose domain transfer approaches
from weakly-labeled Web images for action localization and event recognition tasks.

\section{Conclusion}
\label{sec:conclusion}

In this chapter we tried  to provide an overview of different solutions for 
visual domain adaptation, including both shallow and deep methods.  
We  grouped the methods both by their similarity concerning the problem (homogeneous \vs heterogeneous data, 
unsupervised \vs semi-supervised scenario)
and  the solutions proposed (feature transformation, instance reweighing, deep models, online learning, \etc.). 
We also reviewed methods that solve DA in the case of  heterogeneous data
as well as approaches that address computer vision problems beyond the image classification, 
such as object detection or multi-object tracking.  Finally, we positioned
 domain adaptation within a larger context by linking it to other
 transfer learning techniques as well 
as to traditional machine learning approaches. 

Due to the lack of the space and the large amount of methods  mentioned, we could only briefly depict each method;  
the interested reader can follow the reference for deeper reading. We also
decided  not to  provide  any comparative experimental  results between these methods for the 
following reasons: (1) Even if many DA methods were tested on the benchmark OFF31 \cite{SaenkoECCV10Adapting} and 
OC10 \cite{GongCVPR12Geodesic} datasets,  papers use often different experimental protocols 
(sampling the source \vs using the whole data, 
unsupervised \vs supervised) 
and different parameter tuning strategies (fix parameter sets,  tuning on the source, cross validation or unknown). 
(2) Results reported in different papers given the same methods (\eg GFK, TCA, SA)  
vary also a lot between different re-implementations. For all these reasons,  
making a fair comparison between all the methods based only on the literature review is rather difficult. 
(3) These datasets are rather small, some methods have published results only with the 
outdated SURFBOV features  and  relying only on these results  is not sufficient 
to derive general conclusions about the  methods. 
For a fair comparison, deep methods  should be compared to shallow methods using deep features
extracted from similar architectures, but both  features extracted from the latest 
deep models and deep DA architectures  build on these models  perform extremely well on OFF31 and OC10 
even without adaptation. 

Most DA solutions in the literature are tested on  these relatively small datasets 
(both in terms of number of classes and number of images). However,  with the proliferation of sensors, 
large amount of heterogeneous data is collected, and hence there is a real  need for solutions being 
able to efficiently exploit them.  This shows a real need for more challenging  datasets 
to evaluate and compare the performance of different methods. The few new DA datasets, 
such as the Testbed  cross-dataset (TB) \cite{TommasiTASKCV14Testbed}
or datasets built for model adaptation  between  photos, paintings and
sketches \cite{KlareICB12STowards,CrowleyCVAA14Search,CastrejonCVPR16Learning,SaxenaTASKCV16Heterogeneous}
while more challenging than the popular OFF31 \cite{SaenkoECCV10Adapting}, OC10 \cite{GongCVPR12Geodesic} 
or MNIST \cite{LecunPIEEE98Gradient} \vs SVHN \cite{NetzerNIPS11Reading}, they 
are only sparsely used. Moreover, except the cross-modal Place dataset \cite{CastrejonCVPR16Learning}, they 
are still small scale and single modality datasets.

We have also seen that  only relatively few papers address adaptation beyond recognition and detection.  
Image and video understanding,   semantic and instance level segmentation, human pose, event and action recognition, motion and 3D scene understanding, 
where trying to simply describe the problem with a vectorial representation 
 and classical domain adaptation,  even when it is possible, has serious limitations. 
 Recently,  these  challenging problems are addressed 
 with deep methods  requiring large amount of labeled data. How to adapt these new models 
between domains with no or very limited amount of data is probably one of the main challenge 
that should be addressed by the visual domain adaptation and transfer learning community in the next few years.


\end{document}